\documentclass[letterpaper]{article} 
\usepackage{aaai23}  
\usepackage{times}  
\usepackage{helvet}  
\usepackage{courier}  
\usepackage[hyphens]{url}  
\usepackage{graphicx} 
\usepackage{natbib}  
\usepackage{algorithm}
\usepackage{algorithmic}
\usepackage[utf8]{inputenc} 
\usepackage[T1]{fontenc}   
\usepackage{url}            
\usepackage{booktabs}       
\usepackage{amsfonts}       
\usepackage{nicefrac}       
\usepackage{microtype}      
\usepackage{xcolor}        
\usepackage{times}
\usepackage{soul}
\usepackage[utf8]{inputenc}
\usepackage{amsmath}
\usepackage{amsthm}
\usepackage{algorithm}
\usepackage{algorithmic}
\usepackage{graphicx}
\usepackage{amsmath}
\usepackage{amssymb}
\usepackage{booktabs}
\usepackage{multirow}
\usepackage{bbding}
\usepackage{overpic}
\usepackage{subfigure}
\usepackage{makecell}
\usepackage{colortbl,xcolor}
\definecolor{Gray}{gray}{0.8}

\usepackage[utf8]{inputenc} 
\usepackage[T1]{fontenc}    
\usepackage{url}            
\usepackage{booktabs}       
\usepackage{amsfonts}       
\usepackage{nicefrac}       
\usepackage{microtype}      
\usepackage{xcolor}         

\usepackage[colorlinks, linkcolor=black,anchorcolor=blue,citecolor=black]{hyperref}
\usepackage{times}
\usepackage{soul}
\usepackage{caption}

\usepackage{newfloat}
\usepackage{listings}
\makeatletter
\newcommand{\printfnsymbol}[1]{%
  \textsuperscript{\@fnsymbol{#1}}%
}
\makeatother
\DeclareCaptionStyle{ruled}{labelfont=normalfont,labelsep=colon,strut=off} 
\floatstyle{ruled}
\newfloat{listing}{tb}{lst}{}
\floatname{listing}{Listing}
\title{Ultra-High-Definition Low-Light Image Enhancement: A Benchmark and Transformer-Based Method}
\author{   
    Tao Wang\textsuperscript{\rm 1},
    Kaihao Zhang\textsuperscript{\rm 2},
    Tianrun Shen\textsuperscript{\rm 1},
    Wenhan Luo\textsuperscript{\rm 3}\thanks{Corresponding authors.},
    Bjorn Stenger\textsuperscript{\rm 4},
    Tong Lu\textsuperscript{\rm 1}\printfnsymbol{1}
}
\affiliations{
    \textsuperscript{\rm 1}State Key Lab for Novel Software Technology, Nanjing University, China\\
    
    \textsuperscript{\rm 2} Australian National University, Australia\\
    \textsuperscript{\rm 3} Shenzhen Campus of Sun Yat-sen University, China\\
    \textsuperscript{\rm 4} Rakuten Institute of Technology, Japan\\
    \{taowangzj, super.khzhang, whluo.china\}@gmail.com, tiruns@yeah.net, bjorn@cantab.net, lutong@nju.edu.cn
}

\usepackage{bibentry}
\begin{document}

\maketitle
\def\swone{0.98\linewidth}
\def\swonepointfive{0.735\linewidth}
\def\swtwo{0.46\linewidth}
\def\swthree{0.33\linewidth}
\def\swfour{0.25\linewidth}
\def\swfive{0.19\linewidth}
\def\swsix{0.164\linewidth}
\def\swseven{0.14\linewidth}
\def\sweight{0.12\linewidth}
\def\swnine{0.11\linewidth}
\def\swten{0.1\linewidth}
\def\swfigure{0.14\linewidth}

\DeclareRobustCommand\onedot{\futurelet\@let@token\@onedot}
\def\eg{\emph{e.g., }}  
\def\Eg{\emph{E.g}\onedot}
\def\ie{\emph{i.e., }}    \def\Ie{\emph{I.e}\onedot}
\def\cf{\emph{c.f}\onedot}   \def\Cf{\emph{C.f}\onedot}
\def\etc{\emph{etc. }}  \def\vs{\emph{VS}\onedot}
\def\wrt{w.r.t\onedot}            \def\dof{d.o.f\onedot}
\def\etal{\emph{et al.} }
\makeatother

\newcommand{\boxin}[1]{\textcolor{myGreen}{{[Boxin: #1]}}}
\newcommand{\Tref}[1]{Table~\ref{#1}}
\newcommand{\Eref}[1]{Equation~(\ref{#1})}
\newcommand{\Fref}[1]{Figure~\ref{#1}}
\newcommand{\Sref}[1]{Section~\ref{#1}}
\newcommand{\Aref}[1]{Algorithm~\ref{#1}}
\newcommand{\eref}[1]{Eq.~(\ref{#1})}
\newcommand{\fref}[1]{Fig.~\ref{#1}}
\newcommand{\sref}[1]{Sec.~\ref{#1}}
\newcommand{\argmax}{\operatornamewithlimits{argmax}}
\newcommand{\argmin}{\operatornamewithlimits{argmin}}
\newcommand{\bn}{\mathbf{n}}
\newcommand{\bh}{\mathbf{h}}
\newcommand{\bN}{\mathbf{N}}

\mathchardef\mhyphen="2D
\graphicspath{{Figures/}}

\begin{figure*}[t]
\begin{center}
  \begin{overpic}[width=0.85\textwidth]{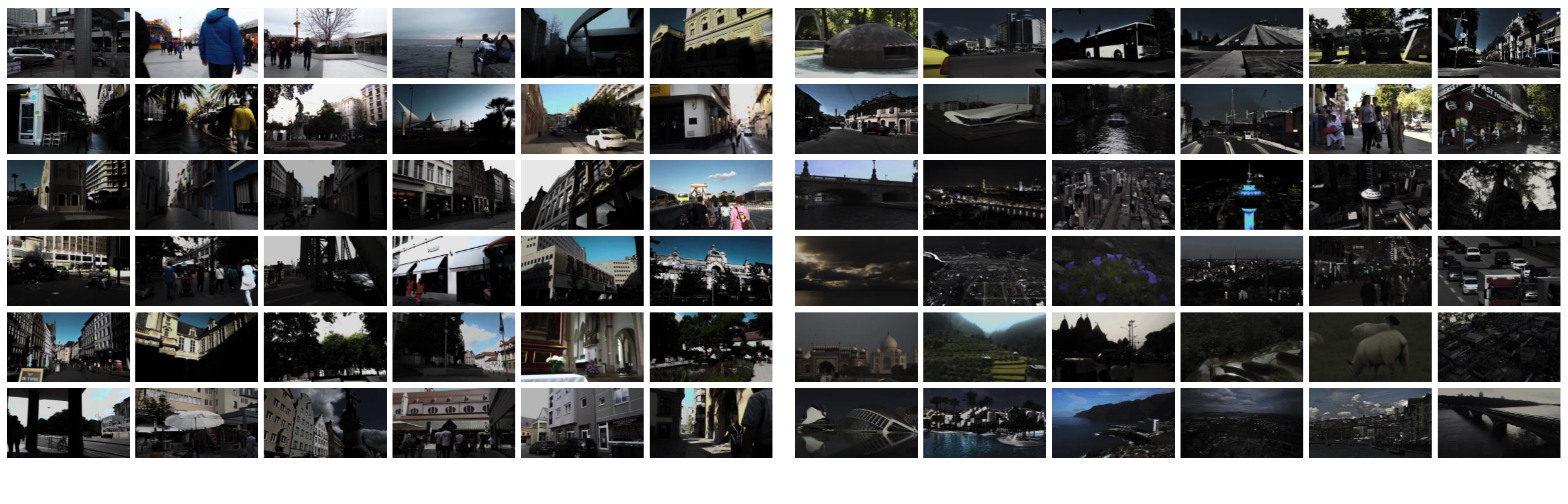} \small
      \put(8,0){(a)~ 4K images from UHD-LOL4K subset.}
  \put(58,0){(b)~ 8K images from UHD-LOL8K subset.}
  \end{overpic}
  \vspace{-0.1in}
    \caption{Low-light images sampled from the proposed UHD-LOL dataset.
  }\label{fig1}
    \vspace{-0.3in}
    \end{center}
\end{figure*}

\begin{abstract}
As the quality of optical sensors improves, there is a need for processing large-scale images. In particular, the ability of devices to capture ultra-high definition (UHD) images and video places new demands on the image processing pipeline. In this paper, we consider the task of low-light image enhancement (LLIE) and introduce a large-scale database consisting of images at 4K and 8K resolution. We conduct systematic benchmarking studies and provide a comparison of current LLIE algorithms. As a second contribution, we introduce LLFormer, a transformer-based low-light enhancement method. The core components of LLFormer are the axis-based multi-head self-attention and cross-layer attention fusion block, which significantly reduces the linear complexity. Extensive experiments on the new dataset and existing public datasets show that LLFormer outperforms state-of-the-art methods. We also show that employing existing LLIE methods trained on our benchmark as a pre-processing step significantly improves the performance of downstream tasks, \eg face detection in low-light conditions. The source code and pre-trained models are available at \url{https://github.com/TaoWangzj/LLFormer}. 
\end{abstract}

\section{Introduction}\label{section:Introduction}
Images taken in low-light conditions typically show noticeable degradation, such as poor visibility, low contrast, and high noise levels. To alleviate these effects, a number of low-light image enhancement (LLIE) methods have been proposed to transform a given low-light image into a high-quality image with appropriate brightness. Traditional LLIE methods are mainly based on image priors or physical models from other tasks, such as histogram equalization-based methods~\cite{kim1997contrast,stark2000adaptive}, retinex-based methods~\cite{kimmel2003variational,wang2014variational} and dehazing-based methods~\cite{dong2011fast,zhang2012enhancement}. Recently, many learning-based LLIE methods have been introduced, making use of large-scale synthetic datasets and achieving significant improvements in terms of performance and speed~\cite{wei2018deep,guo2020zero,lim2020dslr,jiang2021enlightengan,Zero-DCE++,liu2021retinex}.

Most existing datasets, \eg LOL \cite{wei2018deep} and SID \cite{chen2018learning}, consist of lower resolution images (1K or less). Thus, LLIE methods trained on these datasets are naturally constrained to low-resolution images. Sensors on modern mobile devices are able to capture images of resolutions of 4K or 8K, creating a need for algorithms designed for processing Ultra-High-Definition (UHD) images. It is difficult for existing LLIE methods to simultaneously reconcile inference efficiency and visual enhancement on UHD images. In this paper, we focus on the task of Ultra-High Definition Low-Light Image Enhancement (UHD-LLIE). We first build a large-scale benchmark dataset containing UHD images in LOw-Light conditions (UHD-LOL) to explore and evaluate image enhancement algorithms. UHD-LOL includes two subsets, UHD-LOL4K and UHD-LOL8K, containing 4K and 8K-resolution images, respectively. The UHD-LOL4K subset contains $8,099$ image pairs, $5,999$ for training and $2,100$ for testing. The subset of UHD-LOL8K includes $2,966$ image pairs, $2,029$ for training and $937$ for testing. Example 4K and 8K low-light images are shown in Fig.~\ref{fig1}. 

Using this dataset, we conduct extensive benchmarking studies to compare existing LLIE methods and highlight some shortcomings in the UHD setting. We propose a novel Transformer-based method named Low-Light Transformer-based Network (LLFormer) for the UHD-LLIE task. LLFormer is composed of two basic units, an efficient Axis-based Transformer Block and a Cross-layer Attention Fusion Block. Within the Axis-based Transformer block, the axis-based self-attention unit performs the self-attention mechanism on the height and width axes of features across the channel dimension to capture non-local self-similarity and long-range dependencies with less computational complexity. 
Moreover, after the axis-based self-attention, we design a novel Dual Gated Feed-Forward Network, which employs a dual gated mechanism to focus on useful features. The Cross-layer Attention Fusion Block learns attention weights across features in different layers and adaptively fuses features with the learned weights to improve feature representation. The LLFormer adopts a hierarchical structure, which greatly alleviates the computational bottleneck for the UHD-LLIE task. 

To summarize, the contributions of our work are as follows:

$1.$ We build a benchmark dataset of 4K and 8K UHD images, UHD-LOL, to explore and evaluate image enhancement algorithms. To the best of our knowledge, this is the first large-scale UHD low-light image enhancement dataset in the literature.

$2.$ Based on UHD-LOL, we benchmark existing LLIE algorithms to show the performance and limitations of these methods, offering new insights.

$3.$ We propose a novel transformer model, LLFormer, for the UHD-LLIE task. In both quantitative and qualitative aspects, LLFormer achieves state-of-the-art performance on the public LOL and MIT-Adobe FiveK datasets, and our UHD-LOL benchmark.

\section{Related Work}\label{sec:related_work}
\textbf{Low-light Image Datasets.} With the advance in data-driven methods~\cite{zhang2022deep,zhang2020multi}, many datasets for low-light image enhancement have been proposed.
In \cite{vonikakis2013biologically}, Vonikakis \etal introduced an LLIE dataset of $225$ images collected from $15$ different scenes. 
In each scene, they captured $15$ images, including $9$ images under uniform-illumination conditions with different intensities and $6$ images under non-uniform illumination conditions. 
By applying random gamma correction and Gaussian noise, Lore \etal \cite{lore2017llnet} synthesized $422,500$ low-light image patches from $169$ images. 
\cite{shen2017msr} created an LLIE dataset of $10,000$ image pairs, where $8,000$ for training and $2,000$ for testing. 
\cite{chen2018learning} built the See-in-the-Dark (SID) dataset. It contains $5,094$ short-exposure low-light raw images and their corresponding long-exposure ones.
\cite{cai2018learning} synthesized the SICE dataset from $589$ image sequences with multi-exposure image fusion (MEF) or a high dynamic range (HDR) algorithm. 
\cite{wei2018deep} created the LOw-Light (LOL) dataset, which consists of $485$ image pairs for training and $15$ for testing.
Based on the LOL dataset, Liu \etal \cite{liu2021benchmarking} created VE-LOL-L for training and evaluating LLIE methods, which includes $2,100$ images for training and $400$ for evaluation.
MIT-Adobe FiveK \cite{bychkovsky2011learning} consists of $5,000$ images captured of various indoor and outdoor scenes under different lighting conditions.
In this paper, we introduce a UHD low-light image enhancement benchmark of 4K and 8K images, of sufficient size to be used for training deep models and comparing existing methods.

\textbf{Low-light Image Enhancement Methods.} LLIE aims to recover images from underexposed ones taken in low-light conditions~\cite{li2021low}. Over the past decades, many methods have been proposed for the LLIE task, which can be roughly divided into two categories: non-learning based methods and data-driven methods. 
Traditional methods mainly include histogram-based methods~\cite{lee2012contrast,xu2013generalized,celik2014spatial}, Retinex-based methods~\cite{kimmel2003variational,wang2014variational}, and dehazing-based methods~\cite{zhang2012enhancement,li2015low}. 
Although traditional methods obtain reasonable results, the enhanced images of these methods usually suffer from artifacts like color distortion or over enhancement. 
Data-driven methods~\cite{wei2018deep,zhang2020deblurring,lim2020dslr,ni2020towards,zeng2020learning,wang2022low,wang2021video,yang2022adaint} have been successfully applied to the LLIE task. 
For example, RetinexNet in~\cite{wei2018deep} combines Retinex theory and deep CNNs in a unified end-to-end learning framework. DSLR \cite{lim2020dslr} applies a Laplacian pyramid scheme to enhance global and local details in multiple streams of the encoder-decoder architecture. Recently, data-driven methods based on transformers have been applied to low-level tasks: Uformer \cite{wang2021uformer} uses a modified Swin transformer block \cite{liu2021swin} to build a U-shaped network, showing good performance in image restoration. Restormer \cite{zamir2021restormer} introduces modifications of the transformer block for improved feature aggregation for image restoration. While transformers work well in many tasks, their potential for low-light image enhancement remains unexplored. In this work, we focus on designing a transformer for UHD low-light image enhancement.

\section{Benchmark and Methodology}\label{sec:method}

\begin{figure*}[t]
\begin{center}
	\includegraphics[width=0.85\textwidth]{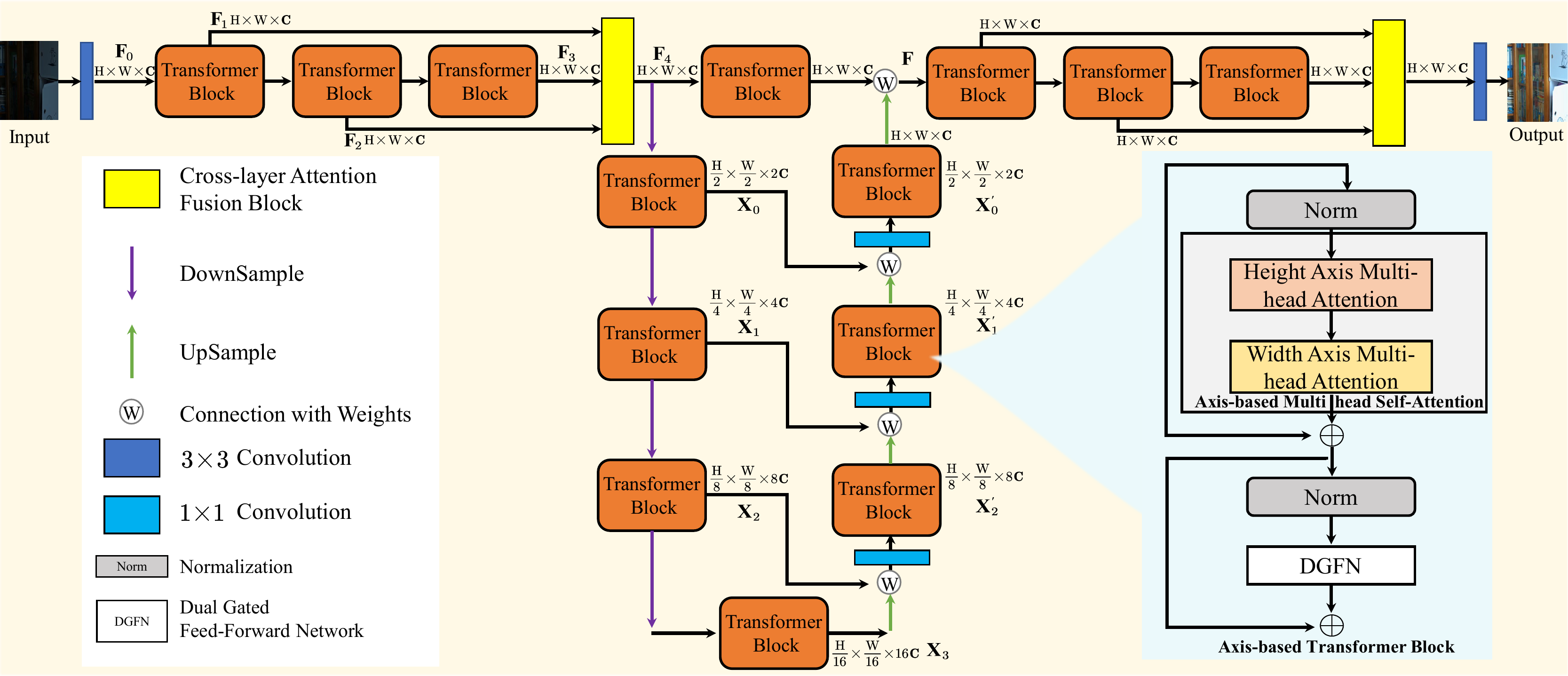}
	  \vspace{-0.1in}
	\caption{{\bf LLFormer architecture}. The core design of LLFormer includes an axis-based transformer block and a cross-layer attention fusion block. In the former, axis-based multi-head self-attention performs self-attention on the height and width axis across the channel dimension sequentially to reduce the computational complexity, and a dual gated feed-forward network employs a gated mechanism to focus more on useful features. The cross-layer attention fusion block learns the attention weights of features in different layers when fusing them.}
	\label{fig:overall}
 \end{center}
 \vspace{-0.3in}
\end{figure*}

\subsection{Benchmark Dataset}
We create a new large-scale UHD-LLIE dataset called UHD-LOL to benchmark the performance of existing LLIE methods and explore the UHD-LLIE problem. UHD-LOL is composed of 4K images of $3,840 \times 2,160$ resolution and 8K images of $7,680 \times 4,320$ resolution, respectively. To build this dataset of image pairs, we use normal-light 4K and 8K images from public data~\cite{zhang2021benchmarking}. These UHD images were crawled from the web and captured by various devices. Images contain both indoor and outdoor scenes, including buildings, streets, people, animals, and natural landscapes. We synthesize corresponding low-light images following~\cite{wei2018deep}, which takes both the low-light degradation process and natural image statistics into consideration. Specifically, we first generate three random variables $X$, $Y$, $Z$, uniformly distributed in $(0,1)$. We use these variables to generate parameters provided by the Adobe Lightroom software. The parameters include exposure ($-5+5X^{2}$), highlights ($50 \min \{Y, 0.5\}+75$), shadows ($-100 \min\{Z,0.5\}$), vibrance ($-75+75X^{2}$), and whites ($16(5-5X^{2})$). The synthesized low-light and normal-light images make up our UHD-LOL, which consists of two subsets: UHD-LOL4K and UHD-LOL8K. The UHD-LOL4K subset contains $8,099$ pairs of 4K low-light/normal-light images. Among them, $5,999$ pairs of images are used for training and $2,100$ for testing. The UHD-LOL8K subset includes $2,966$ pairs of 8K low-light/normal-light images, which are split into $2,029$ pairs for training and $937$ for testing. Example images are shown in Fig.~\ref{fig1}.

\subsection{LLFormer Architecture}
As illustrated in Fig. \ref{fig:overall}, the overall architecture of LLFormer is a hierarchical encoder-decoder structure. Given a low-light image $\mathbf{I} \in \mathbb{R}^{H \times W \times 3}$, LLFormer first employs a $3\times3$ convolution as a projection layer to extract shallow feature $\mathbf{F_{0}} \in \mathbb{R}^{H \times W \times C}$. Next, $\mathbf{F_{0}}$ is fed into three sequential Transformer blocks to extract deeper features. More specifically, intermediate features outputted from transformer blocks are denoted as $\mathbf{F_{1}}, \mathbf{F_{2}}, \mathbf{F_{3}} \in \mathbb{R}^{H \times W \times C}$. These features $\mathbf{F_{1}}, \mathbf{F_{2}}, \mathbf{F_{3}}$ pass through the proposed cross-layer attention fusion block to be aggregated and transformed into the enhanced image features $\mathbf{F_{4}}$.
Second, four stages in an encoder are used for deep feature extraction on $\mathbf{F_{4}}$. To be specific, each stage contains one downsampling layer and multiple transformer blocks. From top to bottom stages, the number of transformer blocks increases.
We use the pixel-unshuffle operation \cite{shi2016real} to downscale the spatial size and double the channel number. Therefore, features in the $i \text {-th}$ stage of the encoder can be denoted as $\mathbf{X}_{i} \in \mathbb{R}^{\frac{H}{2^{i}} \times \frac{W}{2^{i}} \times 2^{i} C}$ and $i = 0,1,2,3$ corresponding to the four stages. Subsequently, the low-resolution latent feature $\mathbf{X}_{3}$ passes through a decoder which contains three stages and takes $\mathbf{X}_{3}$ as input and progressively restores the high-resolution representations. Each stage is composed of an upsampling layer and multiple transformer blocks. Features in the $i \text {-th}$ stage of decoder are denoted as $\mathbf{X}_{i}^{\prime} \in \mathbb{R}^{\frac{H}{2^{i}} \times \frac{H}{2^{i}} \times 2^{i+1} C}, i = 0,1,2$. We apply the pixel-shuffle operation \cite{shi2016real} for upsampling. To alleviate the information loss in the encoder and for features to be well recovered in the decoder, we use the weighted skip connection with a $1 \times 1$ convolution for feature fusion between the encoder and decoder, which can flexibly adjust the contributions of the features from encoder and decoder. Third, after the decoder, the deep feature $\mathbf{F}$ in turn passes through three transformer blocks and a cross-layer attention fusion block to generate the enhanced features for image reconstruction. Finally, LLFormer applies a $3 \times3$ convolution on the enhanced features to yield the enhanced images $\hat{I}$. We optimize LLFormer using a smooth $L_{1}$ loss \cite{girshick2015fast}.

\begin{figure*}[t]
\begin{center}
	\includegraphics[width=0.95\textwidth]{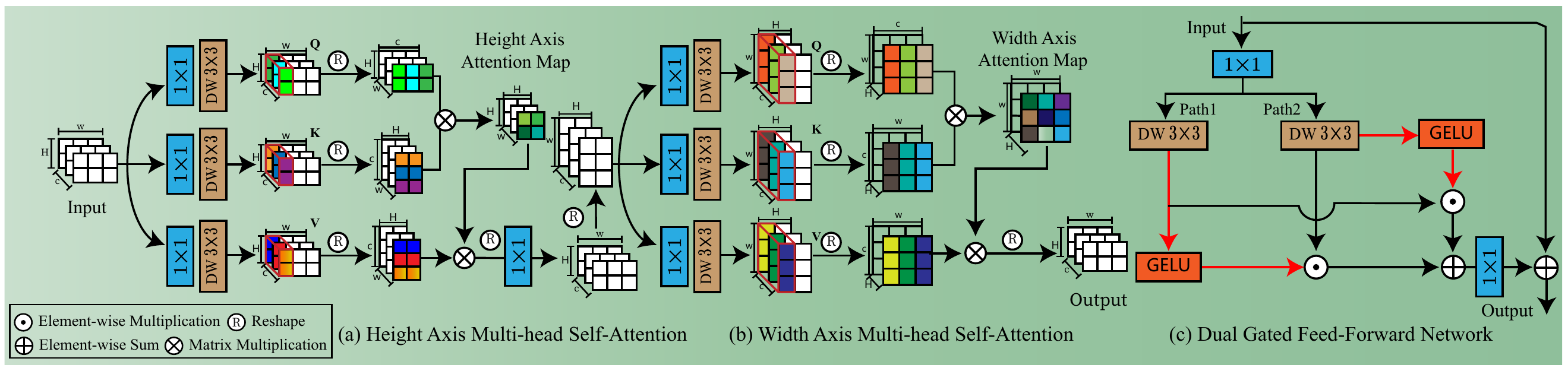}
	  \vspace{-0.1in}
	\caption{The architecture of our Axis-based Multi-head Self-Attention and Dual Gated Feed-Forward Network. From left to right, the components are Height Axis Multi-head Attention, Width Axis Multi-head Attention, and Dual Gated Feed-Forward Network.}
	\label{fig:atten}
 \end{center}
 \vspace{-0.3in}
\end{figure*}

\subsection{Axis-based Transformer Block} \label{sec:self_attention}
Transformers were shown to have advantages in modeling non-local self-similarity and long-range dependencies compared to CNNs. 
However, as discussed in \cite{vaswani2017attention,liu2021swin}, the computational cost of the standard Transformer is quadratic with respect to the spatial size of input feature maps ($H\times W$). Moreover, it often becomes infeasible to apply transformers to high-resolution images especially UHD images. To address this problem, we propose an axis-based multi-head self-attention (A-MSA) mechanism in the transformer block. The computational complexity of A-MSA is linear in spatial size, which greatly reduces the computational complexity. Further, we introduce a dual gated mechanism in the plain transformer feed-forward network and propose the dual gated feed-forward network (DGFN) to capture more important information in features. 
We integrate our A-MSA and DGFN with the plain transformer units to build the Axis-based Transformer Block (ATB). 
As shown in Fig. \ref{fig:overall}, an ATB contains an A-MSA, a DGFN, and two normalization layers. 
The formula of ATB is:
\begin{equation}
\begin{aligned}
&\mathbf{F}^{\prime}=\mathrm{A \mhyphen MSA}\left(\mathrm{LN}\left(\mathbf{\mathbf{F}_{in}}\right)\right)+ \mathbf{\mathbf{F}_{in}}, \\  
&\mathbf{F}_{\text{out}}=\textrm{DGFN}\left(\mathrm{LN}\left(\mathbf{F}^{\prime}\right)\right)+\mathbf{F}^{\prime},
\end{aligned}
\end{equation}
where $\mathbf{\mathbf{F}_{in}}$ denotes the input of ATB. $\mathbf{F}^{\prime}$ and $\mathbf{F}_{\text {out }}$ are the outputs of A-MSA and DGFN, respectively. $\mathrm{LN}$ is the layer normalization \cite{ba2016layer}. In the following, we provide details of A-MSA and DGFN.

\textbf{Axis-based Multi-head Self-Attention}. The computational complexity of the standard self-attention is quadratic with the resolution of input, \ie $\mathcal{O}\left(W^{2} H^{2}\right)$ for $H\times W$ feature maps. 
Instead of computing self-attention globally, we propose A-MSA, as illustrated in Fig. \ref{fig:overall}, to compute self-attention on the height and width axes across the channel dimension sequentially. Thanks to this operation, the complexity of our A-MSA is reduced to linear. Moreover, to alleviate the limitation of transformers in capturing local dependencies, we employ depth-wise convolutions to help A-MSA focus on the local context before computing a feature attention map \cite{zamir2021restormer,wang2021uformer}. Since the mechanisms of height and width axis multi-head self-attention are similar, we thus only introduce the details of height axis multi-head self-attention for ease of illustration.

For height axis multi-head attention, as shown in Fig. \ref{fig:atten} (a), given feature $\mathbf{X} \in \mathbb{R}^{H \times W \times C}$ output from the normalization layer, we at first apply $1 \times 1$ convolutions to enhance the input feature $\mathbf{X}$, and use $3 \times 3$ depth-wise convolutions to obtain features with enriched local information. 
Then, the output features from $3 \times 3$ depth-wise convolutions are query $\mathbf{Q}$, key $\mathbf{K}$, and value $\mathbf{V}$, as $\mathbf{Q}=W_{3\times3}^{Q} W_{1\times1}^{Q} \mathbf{X}$,$\mathbf{K}=W_{3\times3}^{K} W_{1\times1}^{K} \mathbf{X}$ and $\mathbf{V}=W_{3\times3}^{V} W_{1\times1}^{V} \mathbf{X}$, where $W_{1\times 1}$ and $W_{3\times3}$ denote $1 \times 1$ convolution and $3 \times 3$ depth-wise convolution, respectively. 
After that, the query and key are reshaped for conducting dot-product to generate height axis attention map $\mathbf{A} \in \mathbb{R}^{H \times H \times W}$. To achieve multi-head self-attention, we split the reshaped $\hat{\mathbf{Q}}$, $\hat{\mathbf{K}}$ and $\hat{\mathbf{V}}$ into $k$ heads along the feature channel dimension respectively, as
$\hat{\boldsymbol{\mathbf{Q}}}=\left[\hat{q}_{1}, \ldots, \hat{q}_{k}\right]$, $\hat{\boldsymbol{\mathbf{K}}}=\left[\hat{k}_{1}, \ldots, \hat{k}_{k}\right]$, $\hat{\mathbf{V}}=\left[\hat{v}_{1}, \ldots, \hat{v}_{k}\right]$, where the dimension
of each head is $d_{k}= C / k$. The height axis multi-head self-attention for the $j \text {-th}$ head can be formulated as:
\begin{equation}
\operatorname{SA}\left(\hat{q}_{j}, \hat{k}_{j}, \hat{v}_{j}\right)=\hat{v}_{j} \textrm{softmax}\left(\hat{q}_{j}\hat{k}_{j}/{\boldsymbol{\alpha}}\right),
\end{equation}
where $\hat{q}_{j} \in \mathbb{R}^{H \times d_{k} \times W} $, $\hat{k}_{j} \in \mathbb{R}^{ d_{k} \times H \times W}$ and $\hat{v}_{j} \in \mathbb{R}^{d_{k} \times H \times W }$ denote the $j \text {-th}$ head of $\hat{\mathbf{Q}}$, $\hat{\mathbf{K}}$ and $\hat{\mathbf{V}}$, respectively.
$\boldsymbol{\alpha}$ is a scale factor. The output feature $\mathbf{X}^{\prime}$ can be obtained by:
\begin{equation}
\mathbf{X}^{\prime}=W_{1\times1}\mathbf{Concat}_{j=0}^{k}\left(\mathrm{SA}\left(\hat{q}_{j}, \hat{k}_{j}, \hat{v}_{j}\right)\right),
\end{equation}
where $\mathbf{Concat}$ represents the concatenation operation. 
Finally, we reshape $\mathbf{X}^{\prime}$ to obtain the output feature $\mathbf{X}_{out} \in \mathbb{R}^{H \times W \times C}$ of height axis multi-head attention. 
$\mathbf{X}_{out}$ is forwarded to the width axis multi-head attention (see Fig.~\ref{fig:atten} (b)) to compute self-attention along the width axis.

\textbf{Dual Gated Feed-Forward Network}.
Previous work suggests that Feed-Forward Networks (FFN) demonstrate a limitation in capturing local context~\cite{vaswani2017attention,dosovitskiy2020image}. For efficient feature transformations, we introduce a dual gated mechanism and local information enhancement in FFN, and propose a novel dual gated feed-forward network (DGFN). As shown in Fig.~\ref{fig:atten} (c), for the dual gated mechanism, we first apply dual GELU and element-wise product in two parallel paths to filter the less informative features and then fuse useful information from two paths with an element-wise sum. Further, we apply a $1\times1$ convolution ($W_{1\times1}$) and a $3\times3$ depth-wise convolution ($W_{3\times3}$) in each path to enrich the local information. Given $\mathbf{Y} \in \mathbb{R}^{H \times W \times C}$ as input, the complete DGFN is formulated as:
\begin{equation}
\begin{aligned}
\text {DG} &=\phi\left(W_{3\times3}^{1} W_{1\times1}^{1}\mathbf{Y}\right) \odot (W_{3\times3}^{2} W_{1\times1}^{2}\mathbf{Y}) \\
& + (W_{3\times3}^{1} W_{1\times1}^{1}\mathbf{Y}) \odot \phi\left(W_{3\times3}^{2} W_{1\times1}^{2}\mathbf{Y}\right), \\
\hat{\mathbf{Y}} &= W_{1\times1}\text {DG}(\mathbf{Y})+\mathbf{Y},
\end{aligned}
\end{equation}
where $\hat{\mathbf{Y}} \in \mathbb{R}^{H \times W \times C}$ represents the output features, $\text {DG}$ denotes the dual gated mechanism, $\odot$ is the element-wise multiplication operation, and $\phi$ is the GELU activation function.

\begin{figure*}[t]
\vspace{-0.2in}
\begin{center}
	\includegraphics[width=0.95\textwidth]{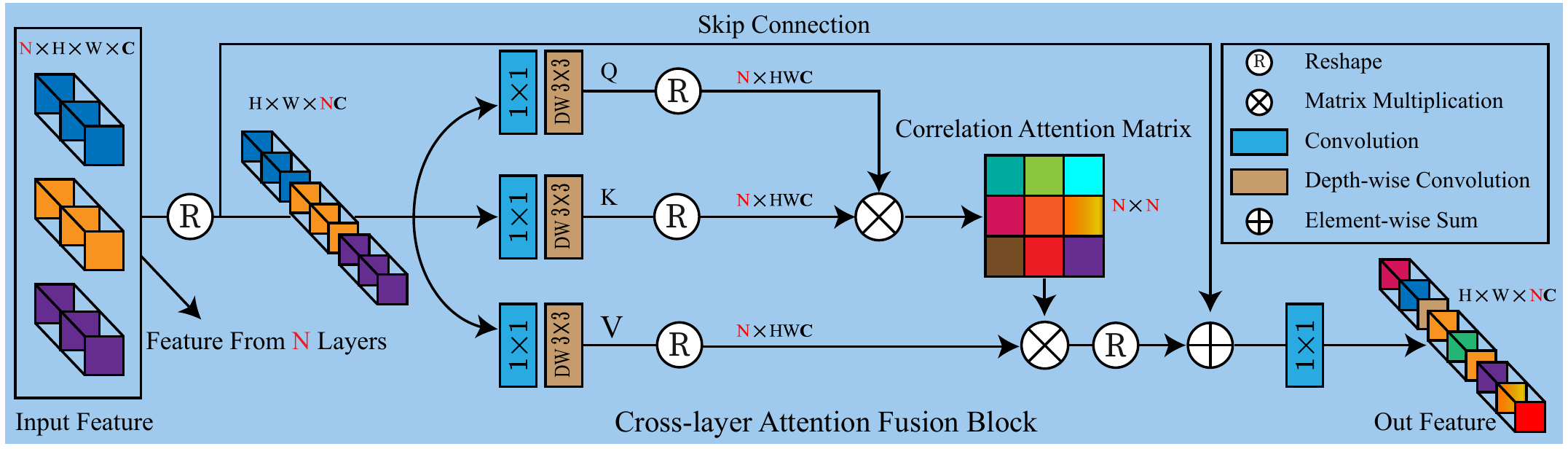}
	  \vspace{-0.1in}
	\caption{The architecture of the proposed Cross-layer Attention Fusion Block. This block efficiently integrates features from different layers with a layer correlation attention matrix.}
	\label{fig:atten2}
 \end{center}
 \vspace{-0.3in}
\end{figure*}

\subsection{Cross-layer Attention Fusion Block }
Recent transformer-based methods adopt feature connections or skip connections to combine features from different layers~\cite{zamir2021restormer,wang2021uformer}.
However, these operations do not fully exploit dependencies across different layers, limiting the representation capability. 
To address this, we propose a novel cross-layer attention fusion block (CAFB), which adaptively fuses hierarchical features with learnable correlations among different layers. 
The intuition behind CAFB is that activations at different layers are a response to a specific class, and feature correlations can be adaptively learned using a self-attention mechanism.

The CAFB architecture is shown in Fig.~\ref{fig:atten2}. Given concatenation features ($\mathbf{F_{in}} \in \mathbb{R}^{N \times H \times W \times C}$) from $N$ successive layers ($N=3$ in the experiments), we first reshape $\mathbf{F_{in}}$ into $\mathbf{\hat{\mathbf{F}}_{in}}$ with dimensions $H \times W \times NC$. Like self-attention in ATB, we employ $1 \times 1$ convolutions to aggregate pixel-wise cross-channel context followed by $3\times3$ depth-wise convolutions to yield $\mathbf{Q}$, $\mathbf{K}$ and  $\mathbf{V}$. We then reshape the query and key into 2D matrices of dimensions $N\times HWC$ ($\hat{\mathbf{Q}}$) and $HWC \times N $ ($\hat{\mathbf{K}}$) to calculate the layer correlation attention matrix $\mathbf{A}$ of size $N \times N$.
Finally, we multiply the reshaped value $\hat{\mathbf{V}} \in \mathbb{R}^{HWC \times N }$ by the attention matrix $\mathbf{A}$ with a scale factor $\boldsymbol{\alpha}$, and add the input features $\mathbf{F_{in}}$. The CAFB process is formulated as:
\begin{equation}
\begin{aligned}
&\mathbf{\hat{\mathbf{F}}_{out}}=W_{1\times1} \textrm{Layer\_Attention }(\hat{\mathbf{Q}}, \hat{\mathbf{K}}, \hat{\mathbf{V}})+\mathbf{\hat{\mathbf{F}}_{in}}, \\
&\textrm{Layer\_Attention }(\hat{\mathbf{Q}}, \hat{\mathbf{K}}, \hat{\mathbf{V}})= \hat{\mathbf{V}} \textrm{softmax}(\hat{\mathbf{Q}}\hat{\mathbf{K}} / \boldsymbol{\alpha}),
\end{aligned}
\end{equation}
where $\mathbf{\hat{\mathbf{F}}_{out}}$ is the output feature that focuses on informative layers of the network. In  practice, we place the proposed CAFB in the symmetric position of the head and tail in the network, so that CAFB helps capture long-distance dependencies among hierarchical layers in both feature extraction and image reconstruction processes. 

\section{Experiments and Analysis}

\subsection{Implementation Details}  
The LLFormer is trained on $128\times128$ patches with a batch size of $12$. For data augmentation, we adopt horizontal and vertical flips. We use the Adam optimizer with an initial learning rate of $10^{-4}$ and decrease it to $10^{-6}$ using cosine annealing. 
The numbers of encoder blocks in the LLFormer from stage $1$ to stage $4$ are $\{2,4,8,16\}$, and the number of attention heads in A-MSA are $\{1,2,4,8\}$. The numbers corresponding to decoders from stage $1$ to $3$ are $\{2,4,8\}$ and $\{1,2,4\}$.
For benchmarking, we compare $16$ representative LLIE methods, including seven traditional non-learning methods (BIMEF~\cite{ying2017bio}, FEA~\cite{dong2011fast}, LIME~\cite{guo2016lime}, MF~\cite{fu2016fusion}, NPE~\cite{wang2013naturalness}, SRIE~\cite{fu2016weighted}, MSRCR~\cite{jobson1997multiscale}), three supervised CNN-based methods (RetinexNet~\cite{wei2018deep}, DSLR~\cite{lim2020dslr}, KinD~\cite{zhang2019kindling}), two unsupervised CNN-based methods (ELGAN~\cite{jiang2021enlightengan}, RUAS~\cite{liu2021retinex}), two zero-shot learning-based methods (Z\_DCE~\cite{guo2020zero}, Z\_DCE++~\cite{Zero-DCE++} ) and two supervised transformer-based methods (Uformer~\cite{wang2021uformer}, Restormer~\cite{zamir2021restormer}). 
For each method, we use the publicly available code and train each learning-based method for $300$ epochs. For ELGAN, we directly use its pre-trained model for testing. Performance is evaluated with the PSNR, SSIM, LPIPS, and MAE metrics.

\begin{table*}[t]\small
\begin{center}
\vspace{-0.1in}
\scalebox{0.75}{\begin{tabular}{l|c|c|c|c||c|c|c|c}  
\hline
\multirow{2}{*}{Methods}  & \multicolumn{4}{c||}{ \textbf{UHD-LOL4K}} & \multicolumn{4}{c}{\textbf{UHD-LOL8K}}  \\ \cline{2-9} 

 & PSNR $\uparrow$ & SSIM $\uparrow$ & LPIPS $\downarrow$ & MAE $\downarrow$  & PSNR $\uparrow$ & SSIM $\uparrow$ & LPIPS $\downarrow$ & MAE $\downarrow$ \\ \hline
    
input images & 11.9439 & 0.5295 & 0.3125&  0.2591& 13.7486  & 0.6415  &  0.3104 & 0.2213    \\ \hline

 BIMEF$^{\dagger}$~\cite{ying2017bio}  & 18.1001 & 0.8876 & 0.1323 & 0.1240  & 19.5225 & 0.9099 & 0.1825 & 0.1048 \\ \cline{2-9} 
 FEA$^{\dagger}$~\cite{dong2011fast}    & 18.3608 & 0.8161 & 0.2197 & 0.0986  & 15.3301 & 0.7699 & 0.3696 & 0.1700\\ \cline{2-9} 
LIME$^{\dagger}$~\cite{guo2016lime} & 16.1709 & 0.8141 & 0.2064 & 0.1285 &  13.5699 & 0.7684 & 0.3055 & 0.2097  \\ \cline{2-9} 
MF$^{\dagger}$~\cite{fu2016fusion}   & 18.8988 & 0.8631 &  0.1358  & 0.1111 &  18.2474 & 0.8781 &  0.2158  & 0.1258 \\ \cline{2-9} 
 NPE$^{\dagger}$~\cite{wang2013naturalness} & 17.6399 &  0.8665 &  0.1753 &  0.1125&  16.2283 & 0.7933 & 0.3214 &  0.1506  \\ \cline{2-9} 
                    
SRIE$^{\dagger}$~\cite{fu2016weighted} & 16.7730 & 0.8365 & 0.1495 & 0.1416 & 19.9637 & 0.9140 & 0.1813 & 0.0975   \\  \cline{2-9} 
                                    
MSRCR$^{\dagger}$~\cite{jobson1997multiscale} &  12.5238 & 0.8106 & 0.2136 & 0.2039 & 12.5238 & 0.7201 & 0.4364& 0.2352 \\ \cline{1-9} 

RetinexNet$^{\ddag}$~\cite{wei2018deep} & 21.6702 & 0.9086 & 0.1478& 0.0690 & 21.2538 &0.9161 & 0.1792&0.0843   \\ \cline{2-9} 
                    
DSLR$^{\ddag}$~\cite{lim2020dslr}  & 27.3361 & 0.9231 & 0.1217& 0.0341 & 21.9406 & 0.8749  & 0.2661 & 0.0805     \\ \cline{2-9} 
                    
KinD$^{\ddag}$~\cite{zhang2019kindling}  & 18.4638 & 0.8863 & 0.1297 & 0.1060 & 17.0200 & 0.7882 & 0.1739&0.1538 \\ \cline{2-9}

Z\_DCE$^{\S}$~\cite{guo2020zero}  & 17.1873 & 0.8498 & 0.1925 &0.1465 &  14.1593 & 0.8141  &0.2847 & 0.1914 \\ \cline{2-9} 
                     
Z\_DCE++$^{\S}$~\cite{Zero-DCE++}  &  15.5793 & 0.8346  & 0.2223  &  0.1701& 14.6837& 0.8348 & 0.2466 & 0.1904     \\ \cline{2-9} 
                     
RUAS$^{\triangle}$~\cite{liu2021retinex} &   14.6806  &  0.7575 & 0.2736  &  0.1690 &  12.2290 & 0.7903 & 0.3557 & 0.2445     \\ \cline{2-9} 
                     
ELGAN$^{\triangle}$~\cite{jiang2021enlightengan}  & 18.3693 & 0.8642  & 0.1967  &  0.1011 &  15.2009&  0.8376 & 0.2293 & 0.1713      \\ \cline{2-9}  
                     
Uformer$^{\star}$~\cite{wang2021uformer}  &\textcolor{purple}{29.9870} & \textcolor{purple}{0.9804} & \textcolor{purple}{0.0342} & \textcolor{purple}{0.0262}& \textcolor{purple}{28.9244}  & \textcolor{purple}{0.9747}  &  \textcolor{purple}{0.0602} & \textcolor{purple}{0.0344} \\ \cline{2-9} 
Restormer$^{\star}$~\cite{zamir2021restormer} & \textcolor{blue}{36.9094}&  \textcolor{blue}{0.9881} &\textcolor{blue}{0.0226} & \textcolor{blue}{0.0117} & \textcolor{blue}{35.0568} & \textcolor{blue}{0.9858} & \textcolor{blue}{0.0331} & \textcolor{blue}{0.0195}  \\ \cline{2-9} 
                   
\textbf{LLFormer$^{\star}$} & \textcolor{red}{37.3340} & \textcolor{red}{0.9889} & \textcolor{red}{0.0200} & \textcolor{red}{0.0116} & \textcolor{red}{35.4313} & \textcolor{red}{0.9861} & \textcolor{red}{0.0267} & \textcolor{red}{0.0194} \\ \hline
\end{tabular}} 
\vspace{-0.1in}
\caption{ Benchmarking study on the UHD-LOL4K and UHD-LOL8K subsets. $\dagger, \ddag, \S, \triangle $ and ${\star}$ indicate the traditional methods, supervised CNN-based methods, unsupervised CNN-based methods, zero-shot methods and transformer-based methods. The top three results are marked in \textcolor{red}{red}, \textcolor{blue}{blue} and \textcolor{purple}{purple}, respectively.}
\label{tab:results1}
\vspace{-0.1in}
\end{center}
\end{table*}

\renewcommand{\tabcolsep}{.5pt}  
\begin{figure*}
\vspace{-0.2cm}
\begin{center}
\begin{tabular}{ccccccccc}
   \vspace{-1.0mm}
   \includegraphics[width=\swnine]{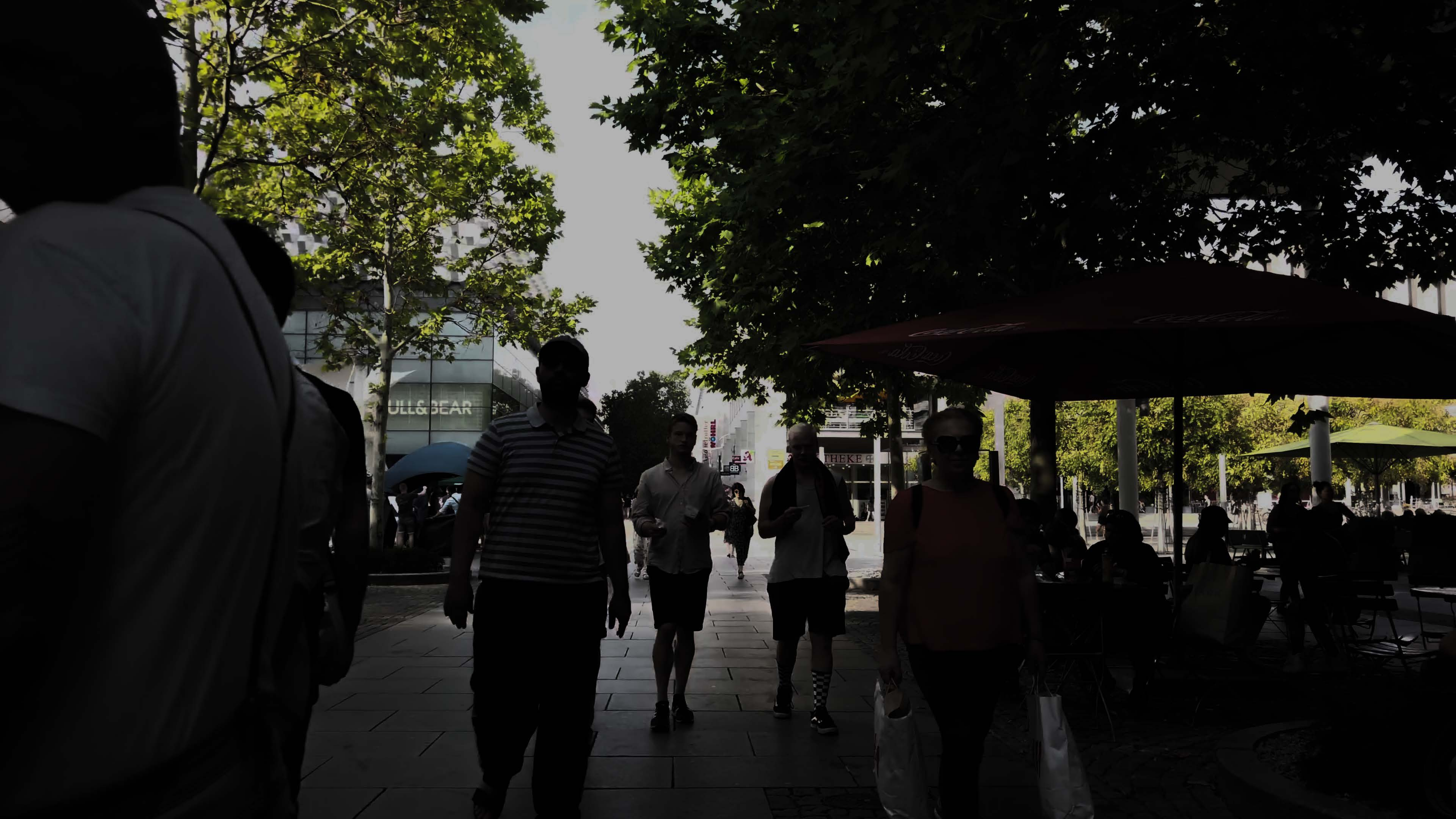}&
   \includegraphics[width=\swnine]{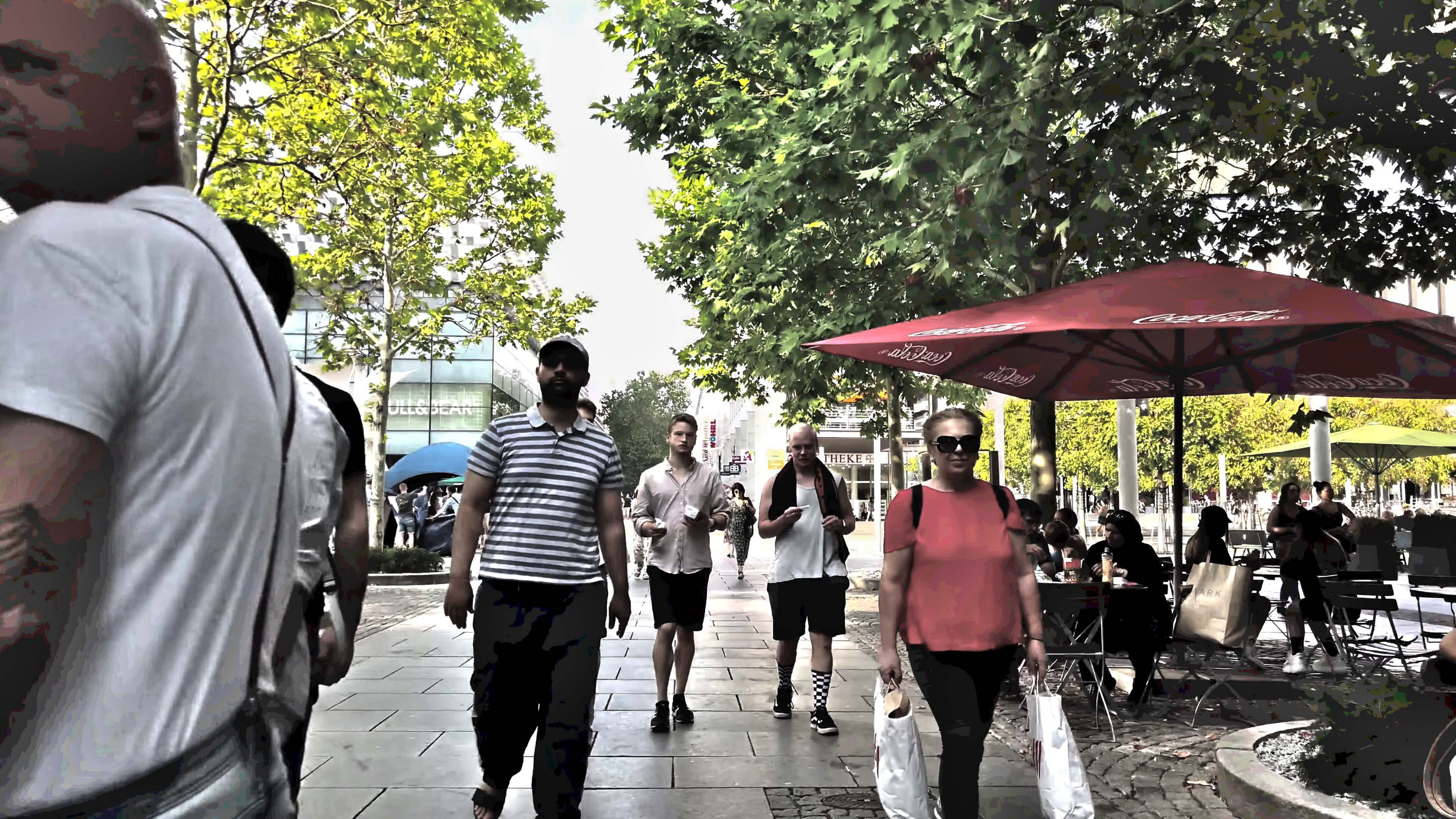}&
   \includegraphics[width=\swnine]{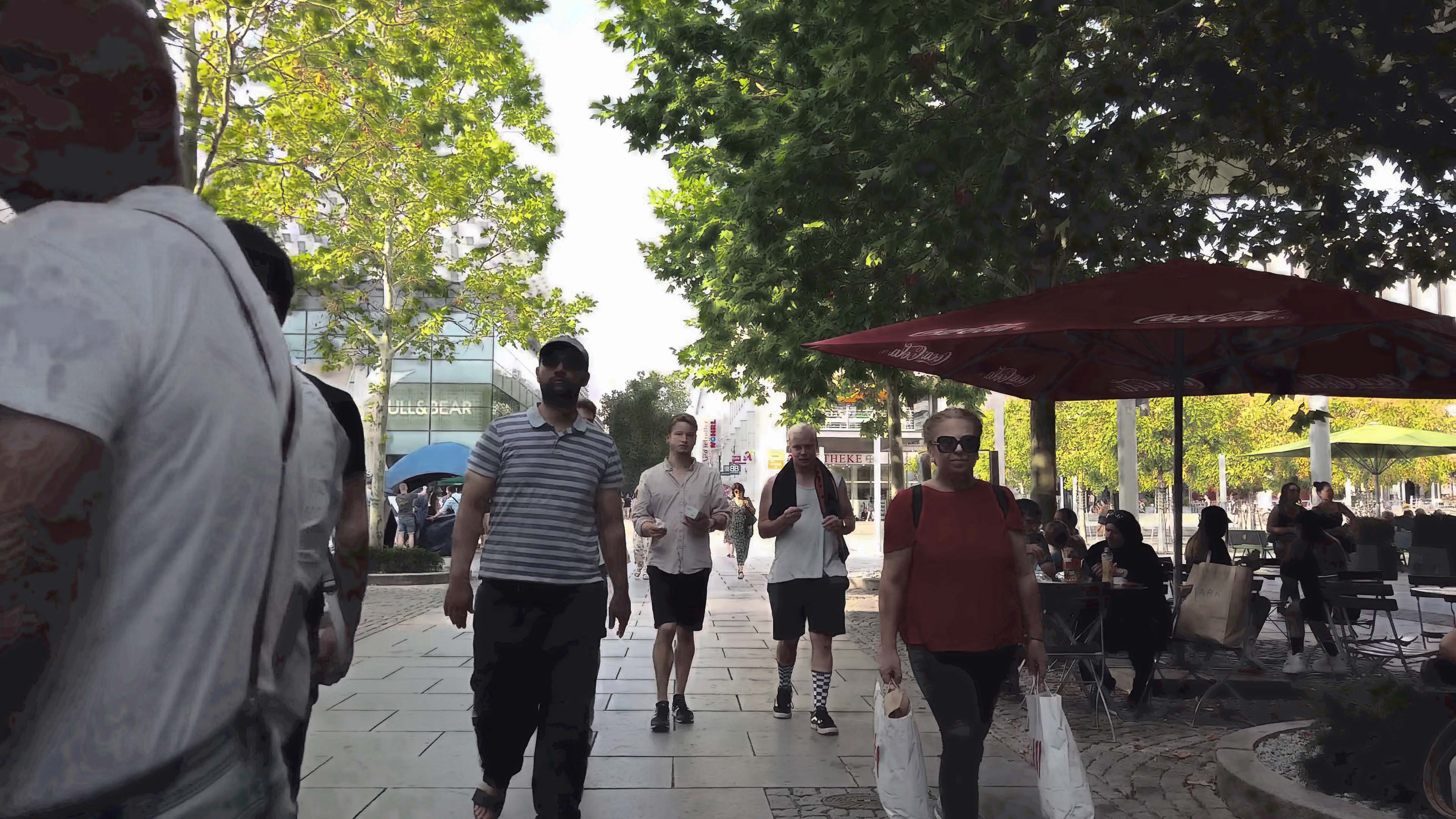}&
   \includegraphics[width=\swnine]{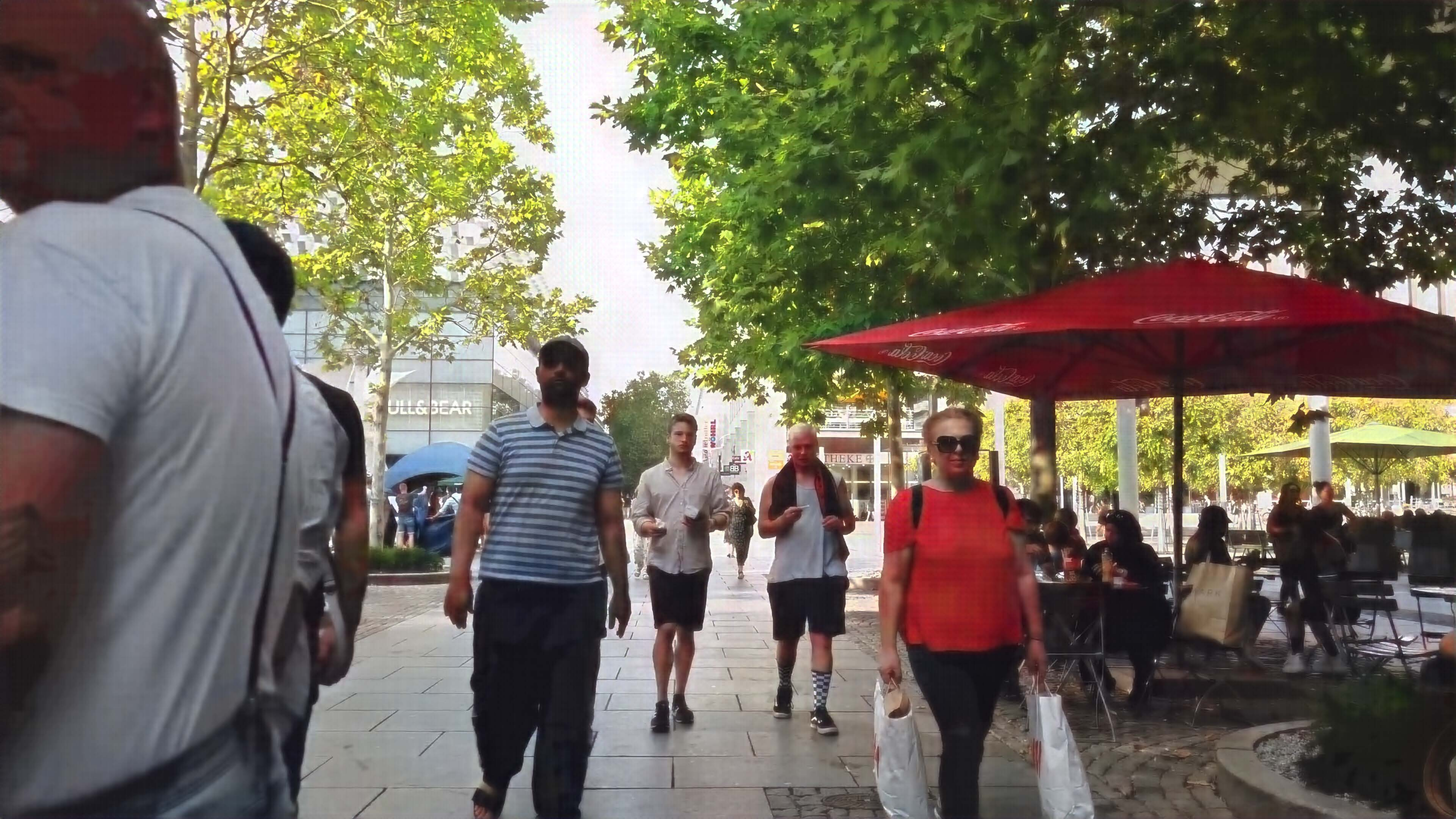}&
   \includegraphics[width=\swnine]{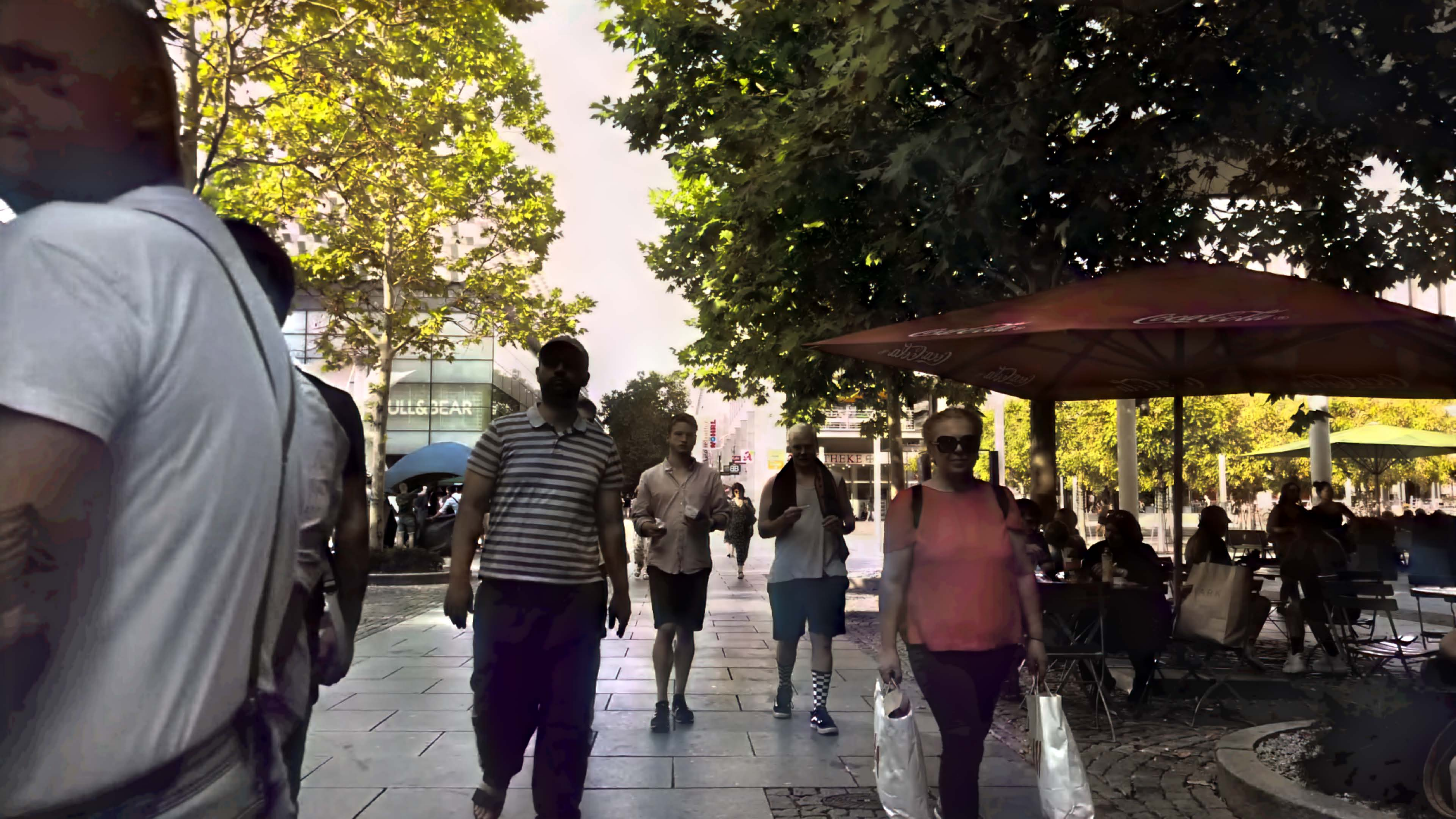} &
   \includegraphics[width=\swnine]{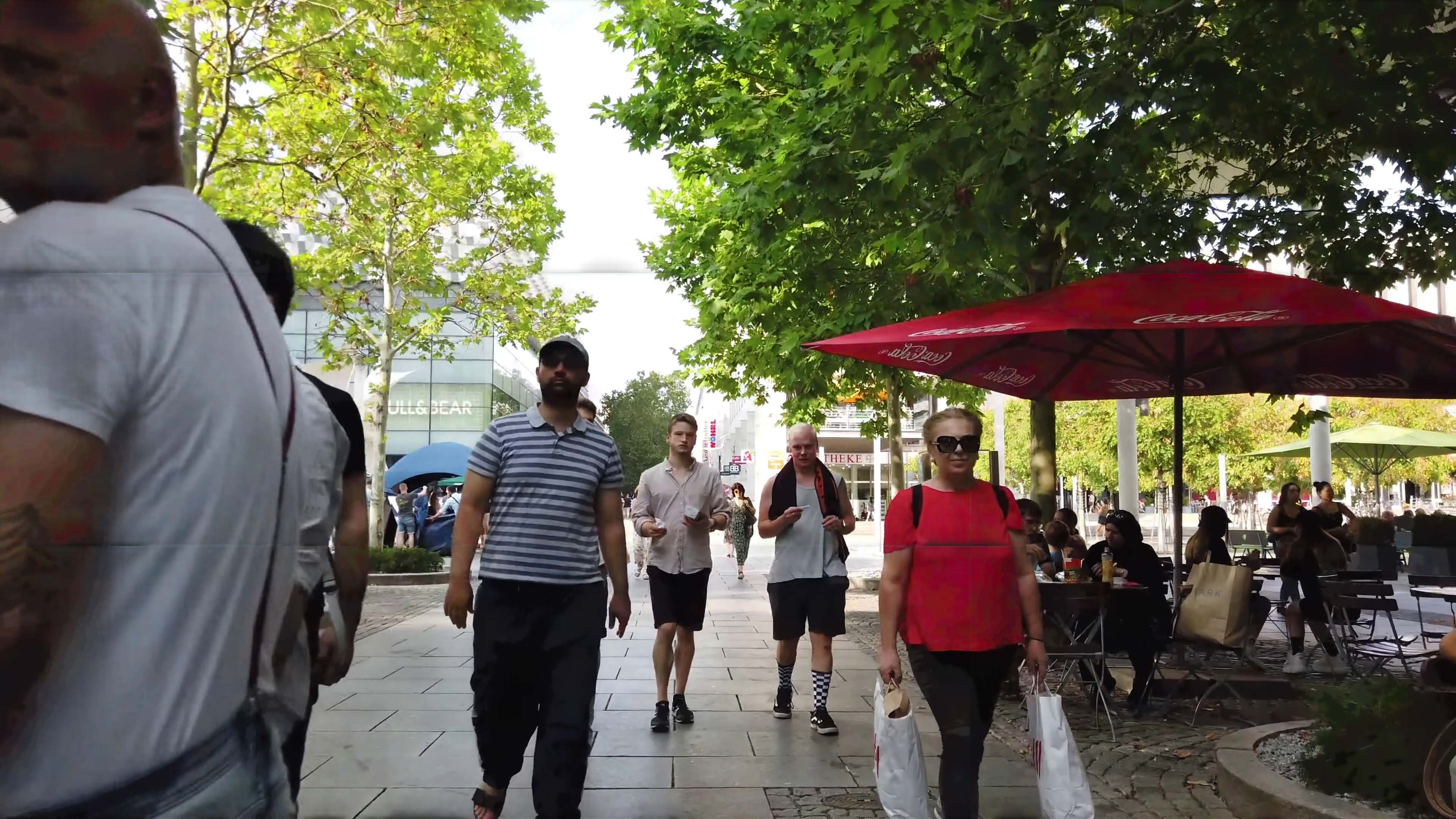}&
   \includegraphics[width=\swnine]{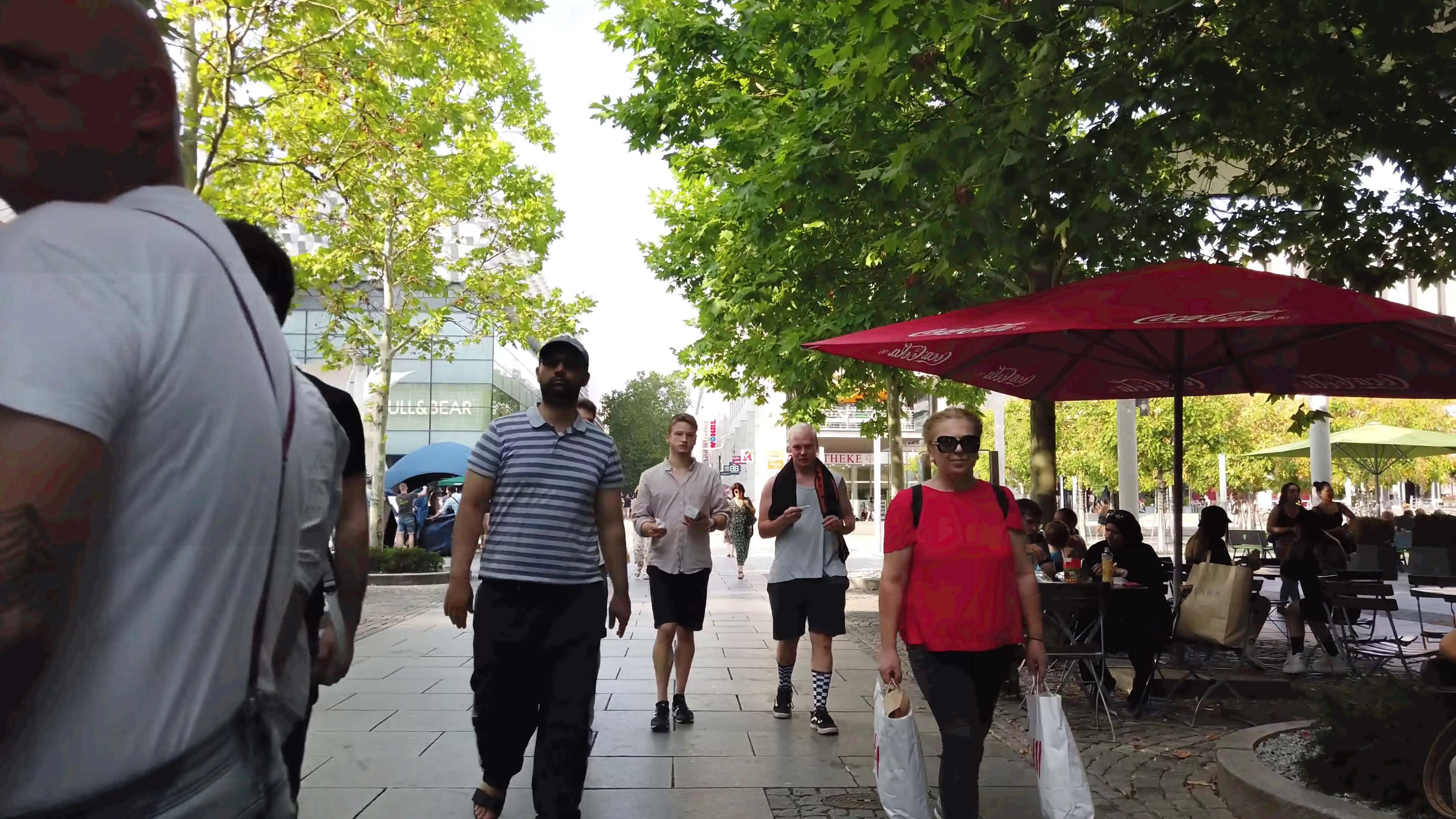}&
   \includegraphics[width=\swnine]{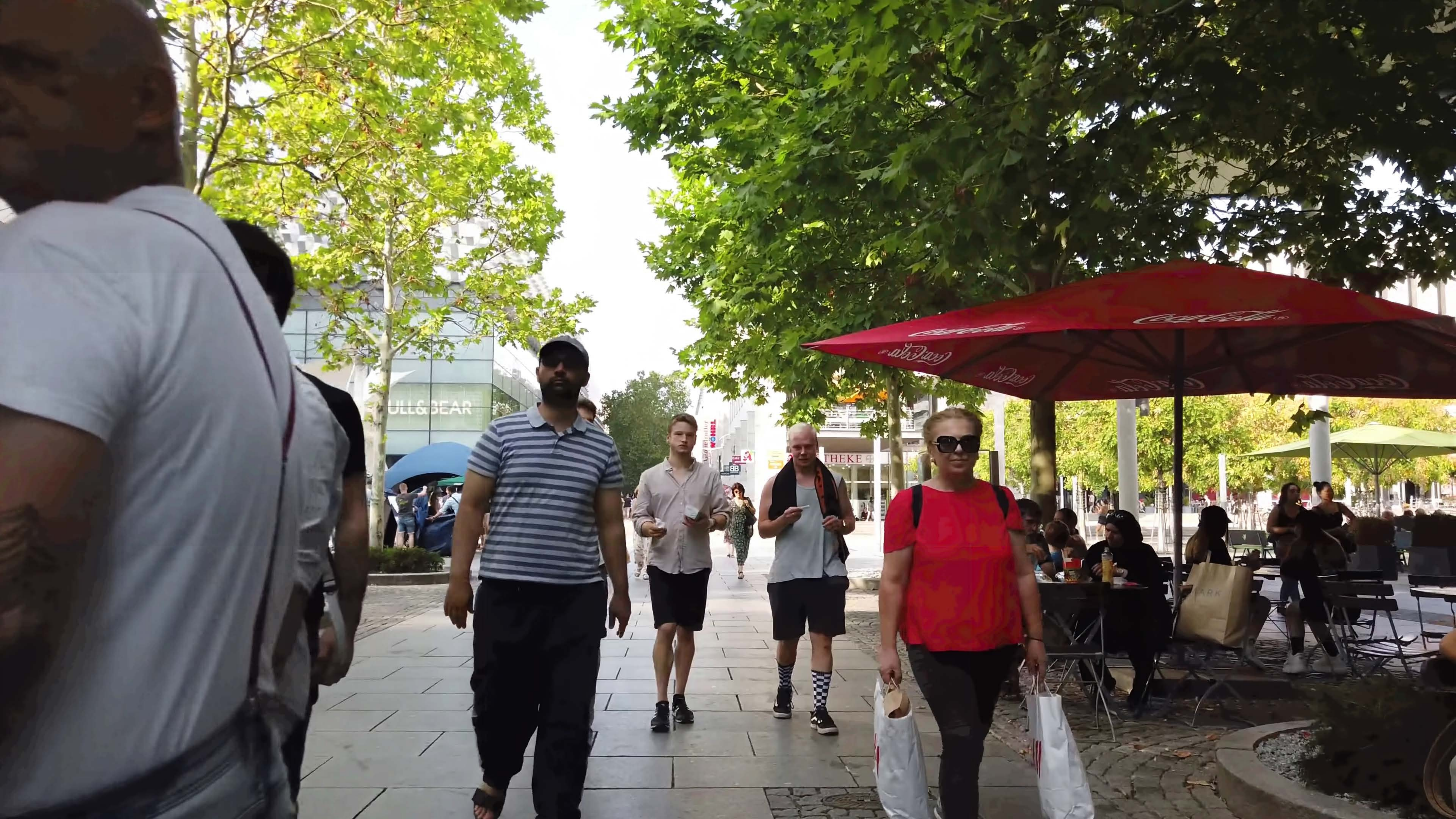}&
   \includegraphics[width=\swnine]{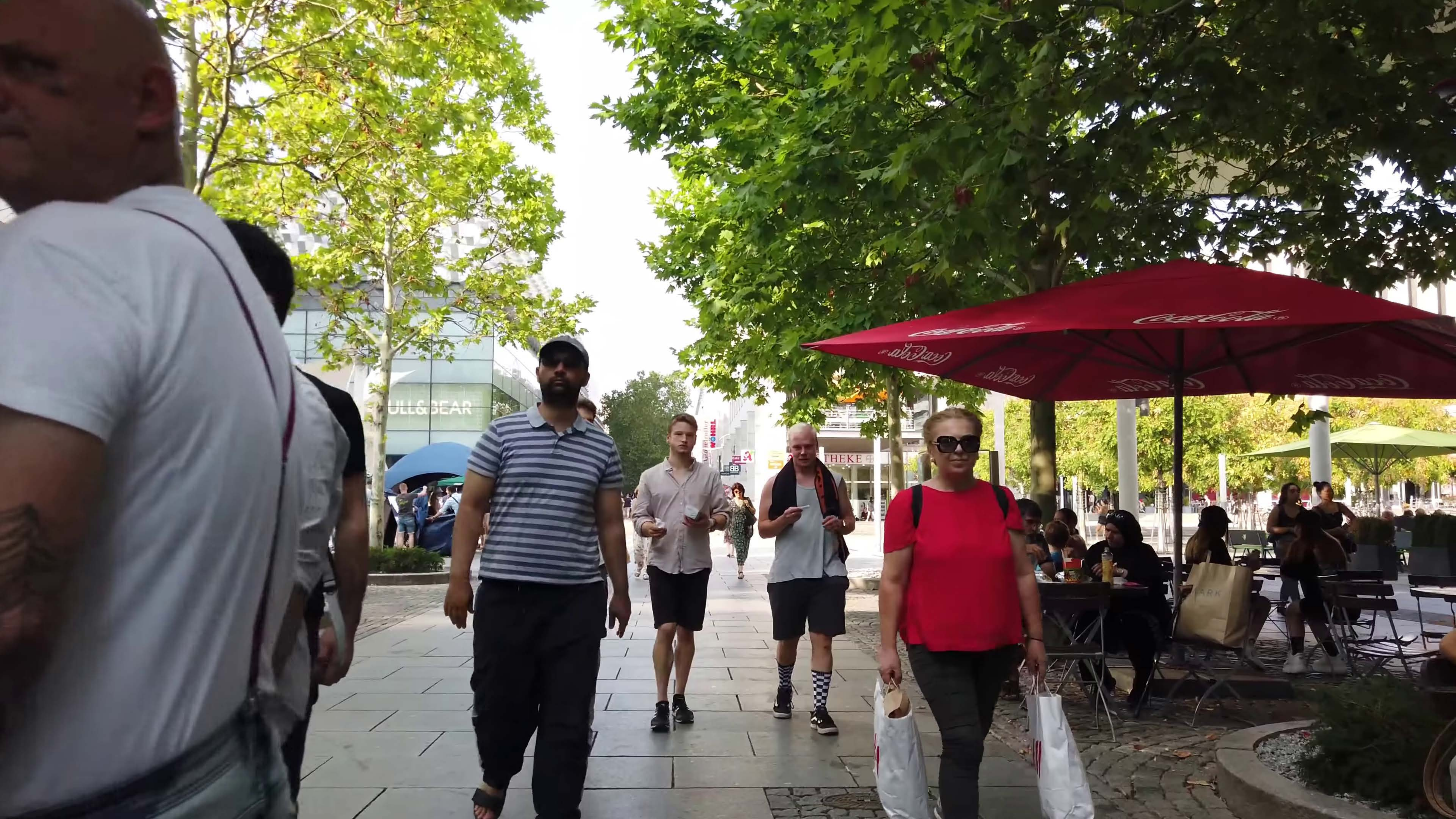}\\
   \vspace{-1.0mm}
   
   \includegraphics[width=\swnine]{Figures/figure5/pdf/057_United_frame00003_INPUT.pdf}&
   \includegraphics[width=\swnine]{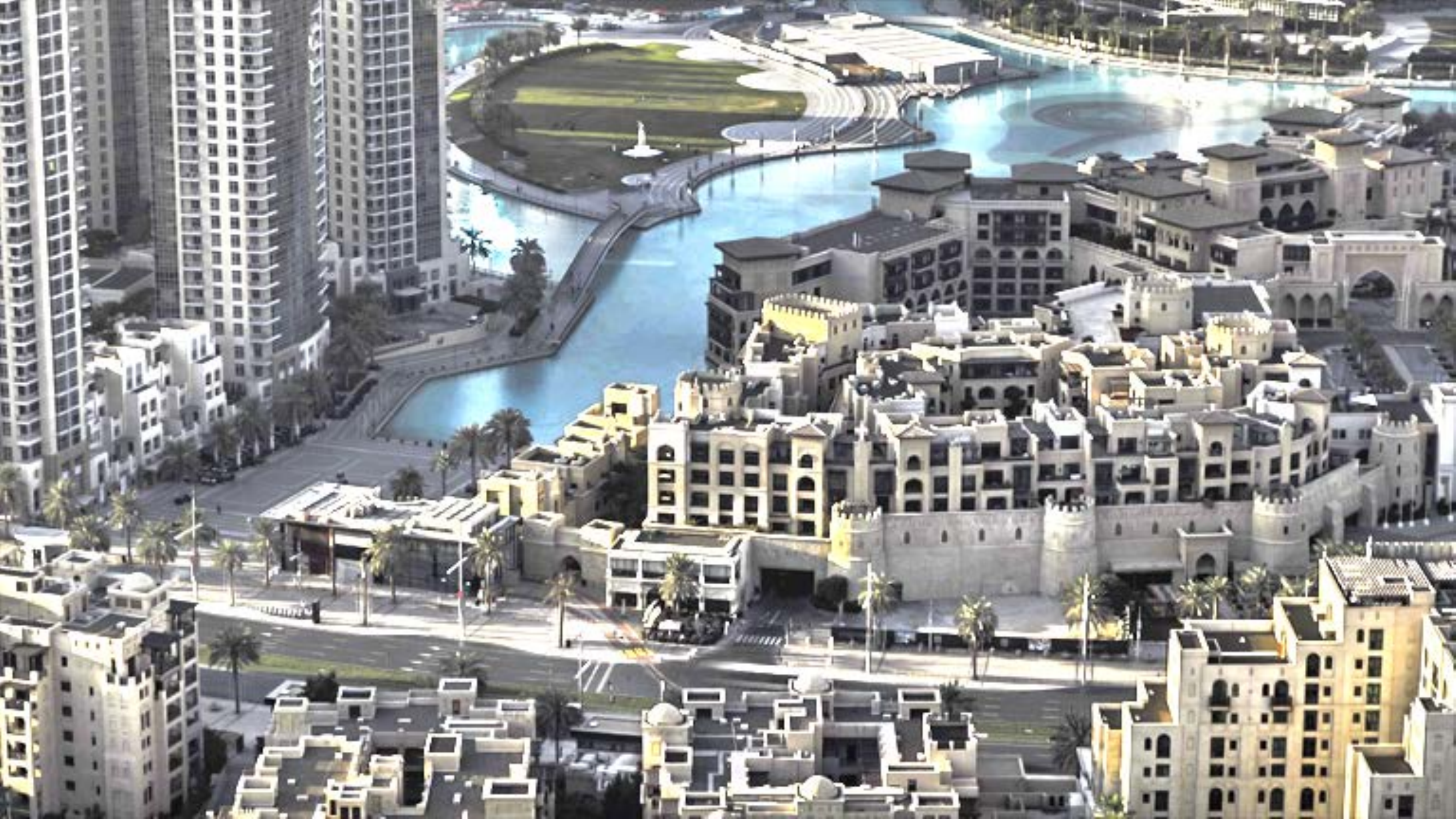}&
   \includegraphics[width=\swnine]{Figures/figure5/pdf/057_United_frame00003_RetinxNet.pdf}&
   \includegraphics[width=\swnine]{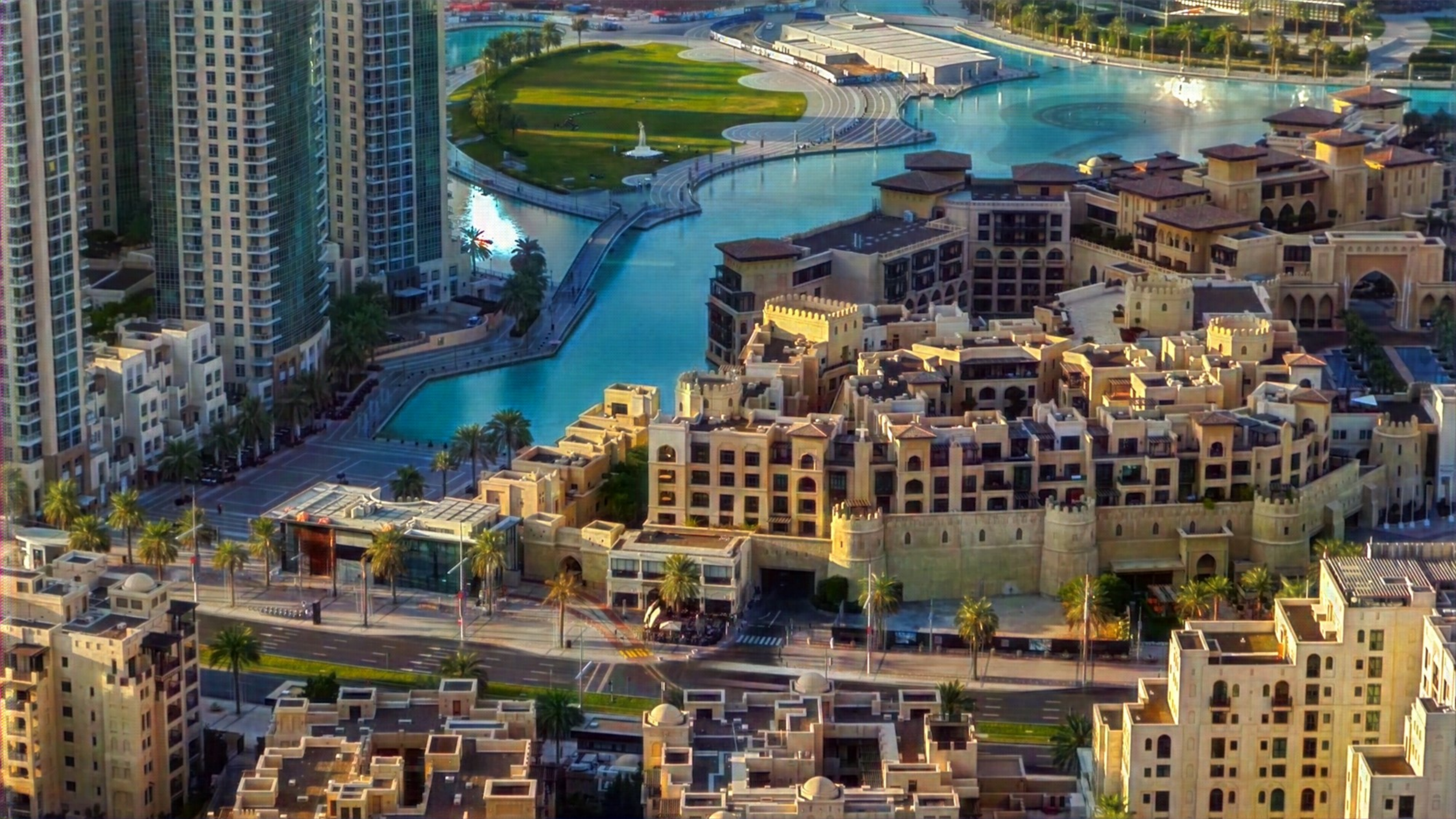}&
   \includegraphics[width=\swnine]{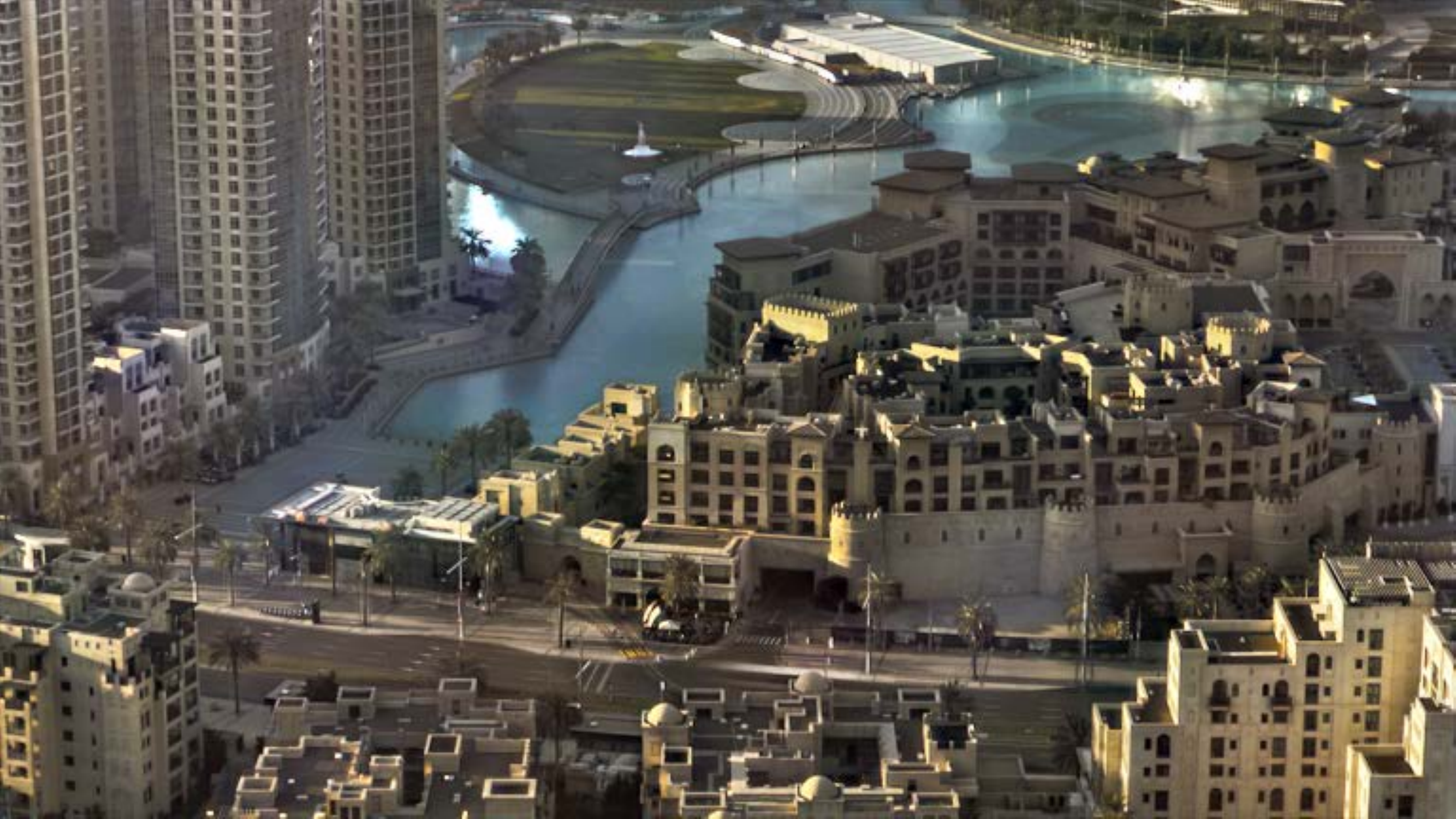} &
   \includegraphics[width=\swnine]{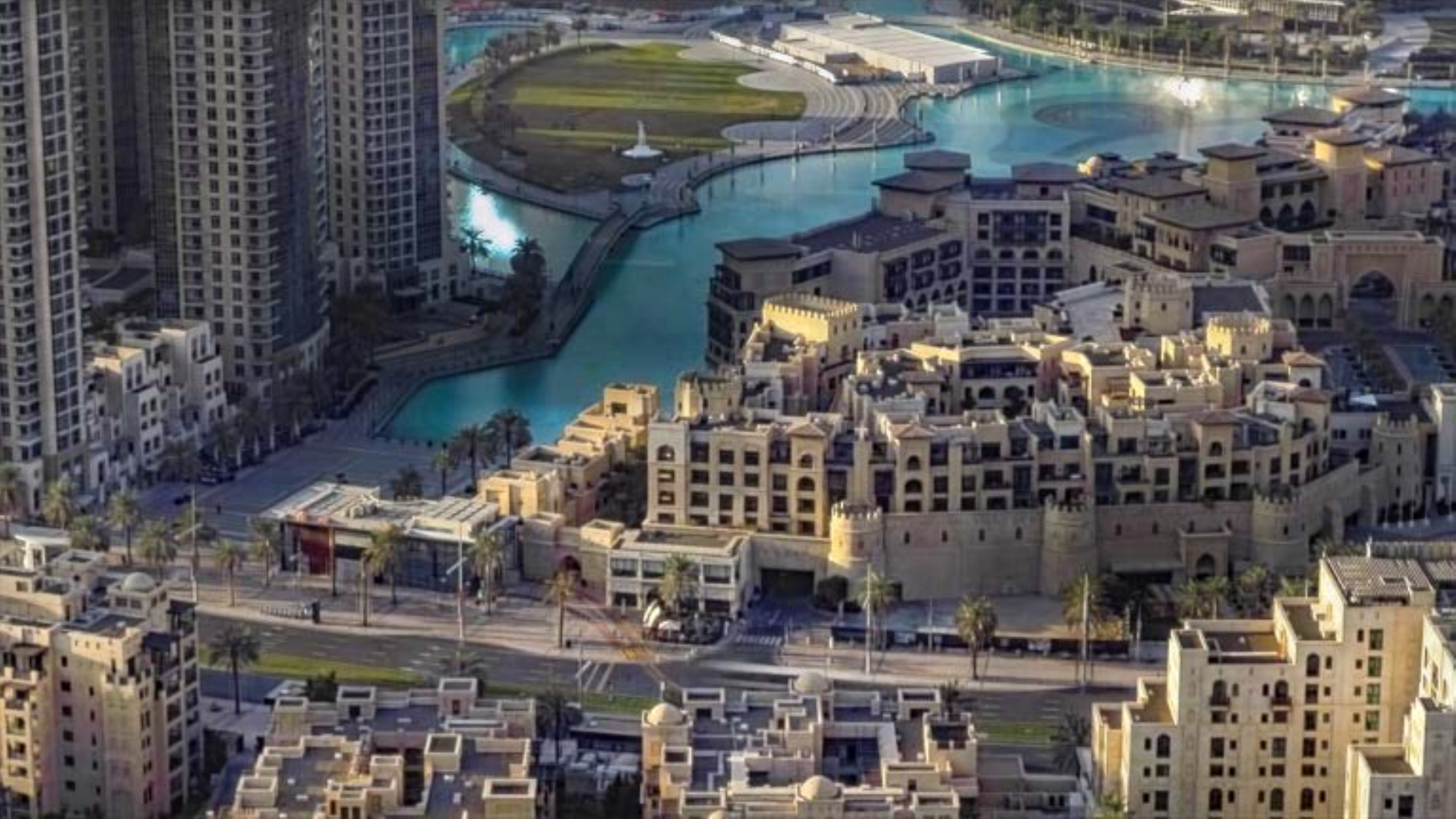}&
   \includegraphics[width=\swnine]{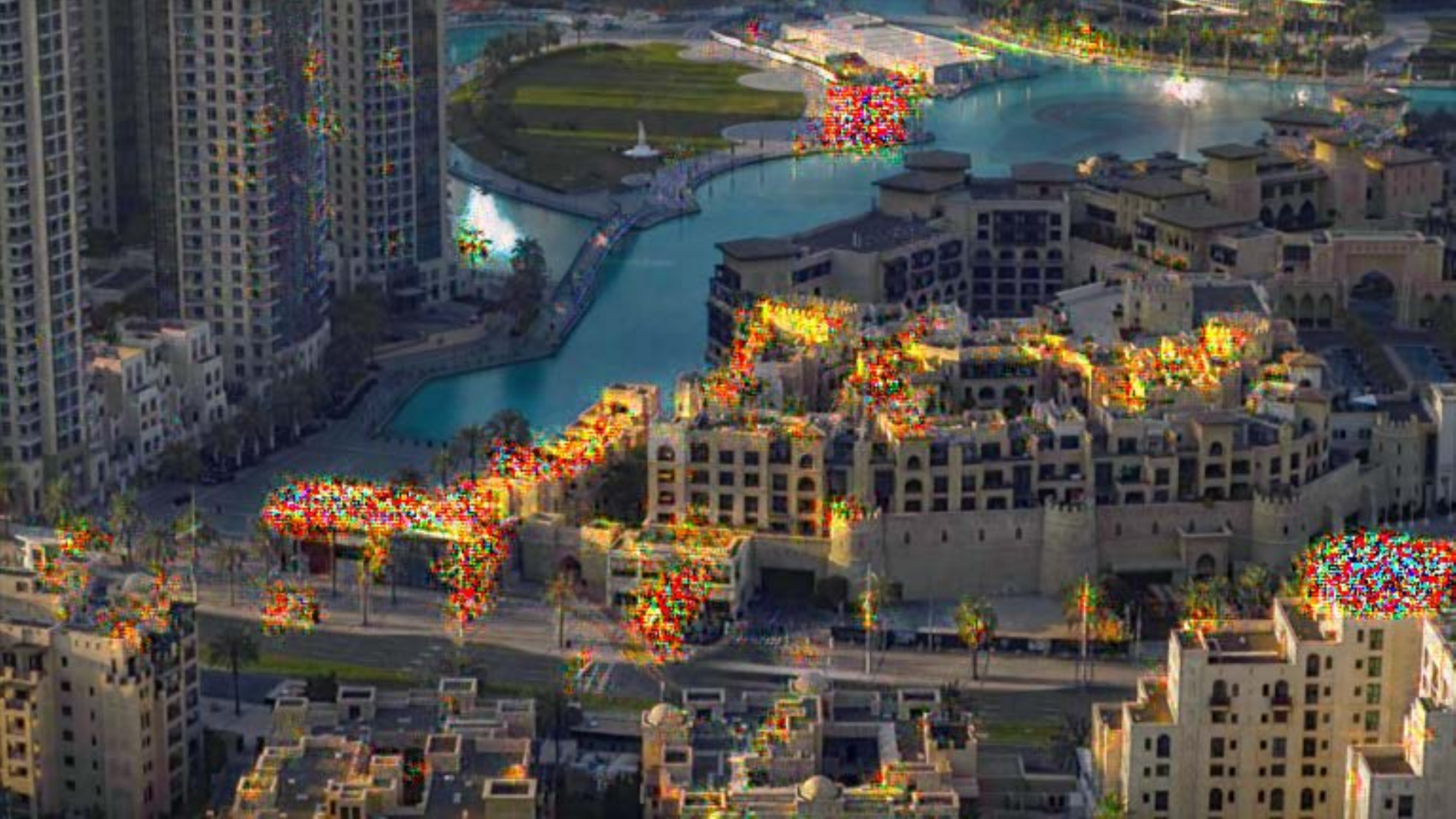}&
   \includegraphics[width=\swnine]{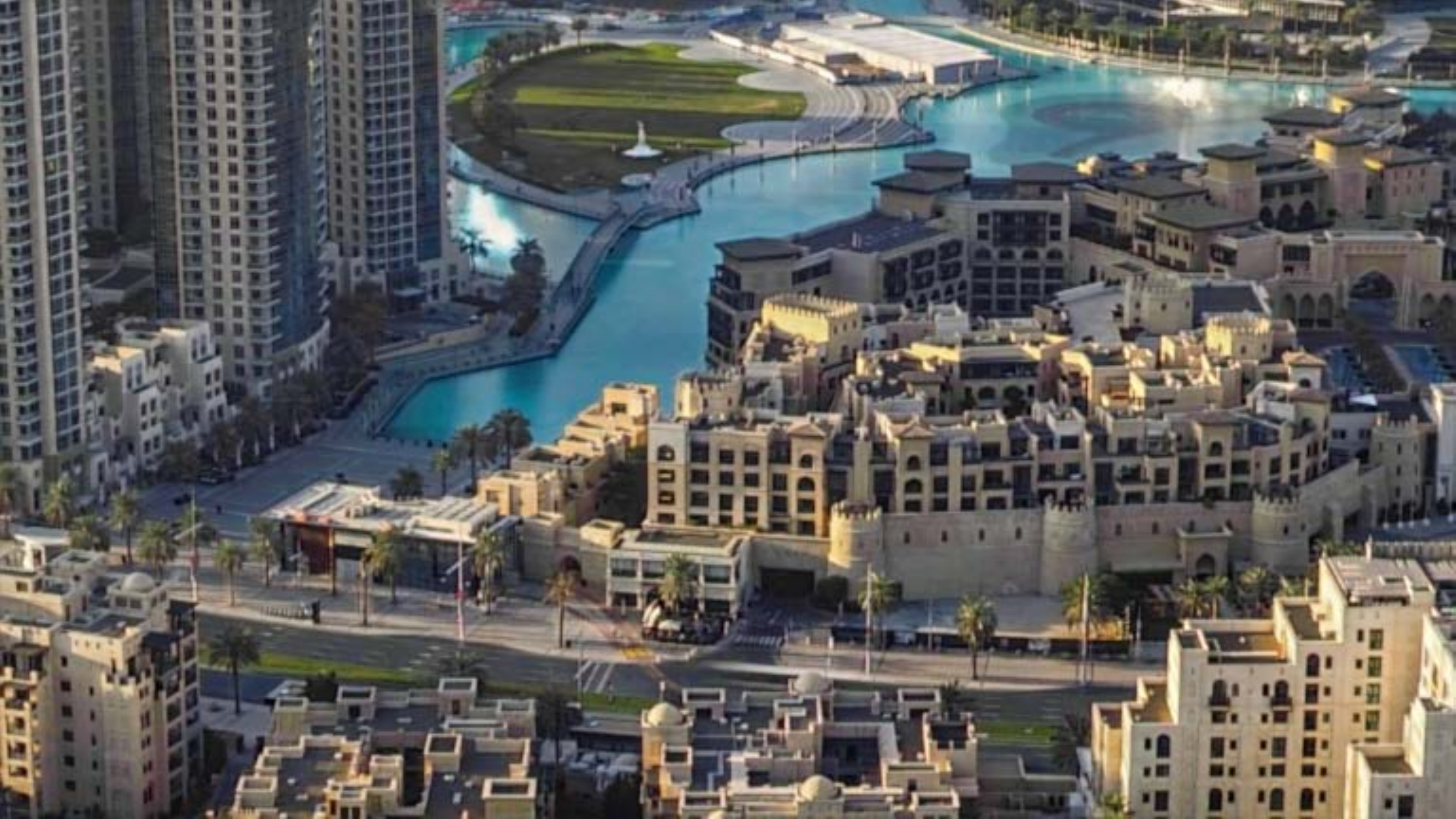}&
   \includegraphics[width=\swnine]{Figures/figure5/pdf/057_United_frame00003_GT.pdf}\\
      
   Input & LIME  & RetinexNet  & DSLR & ELGAN & Uformer & Restormer & \textbf{LLFormer}  & GT
\end{tabular}
\end{center}
\vspace{-5mm}
\caption{Visual results of different methods on the proposed dataset. The top row is from the UHD-LOL4K subset, and the bottom is from the UHD-LOL8K subset. \textbf{Zoom in for a better view}.}
\label{fig:cmp1vAll}
 \vspace{-0.1in}
\end{figure*}%

\renewcommand{\tabcolsep}{.5pt}  
\begin{figure}
\vspace{-0.2cm}
\begin{center}
\scalebox{0.85}{\begin{tabular}{cc}
   \includegraphics[width=\swtwo]{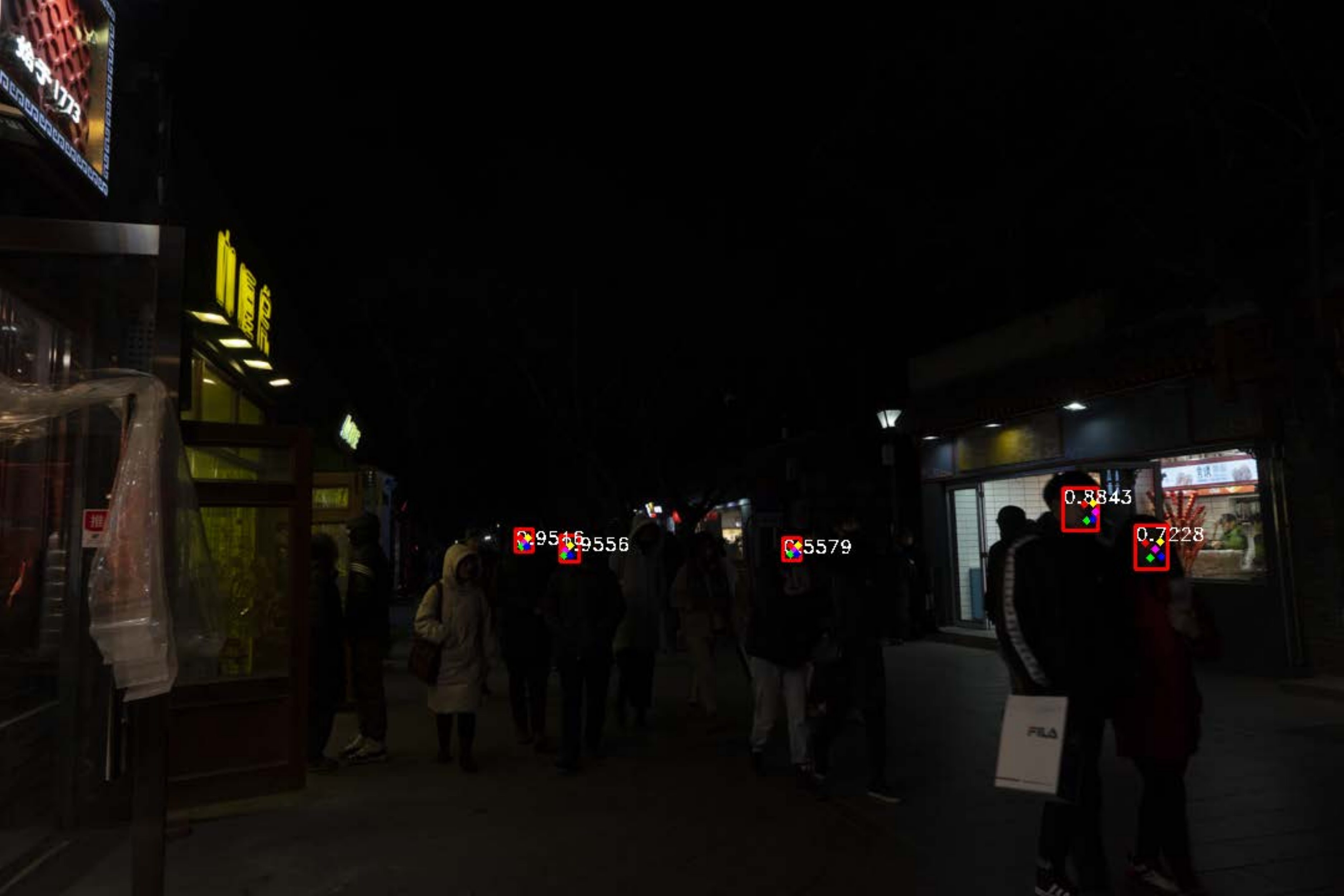}&
   \includegraphics[width=\swtwo]{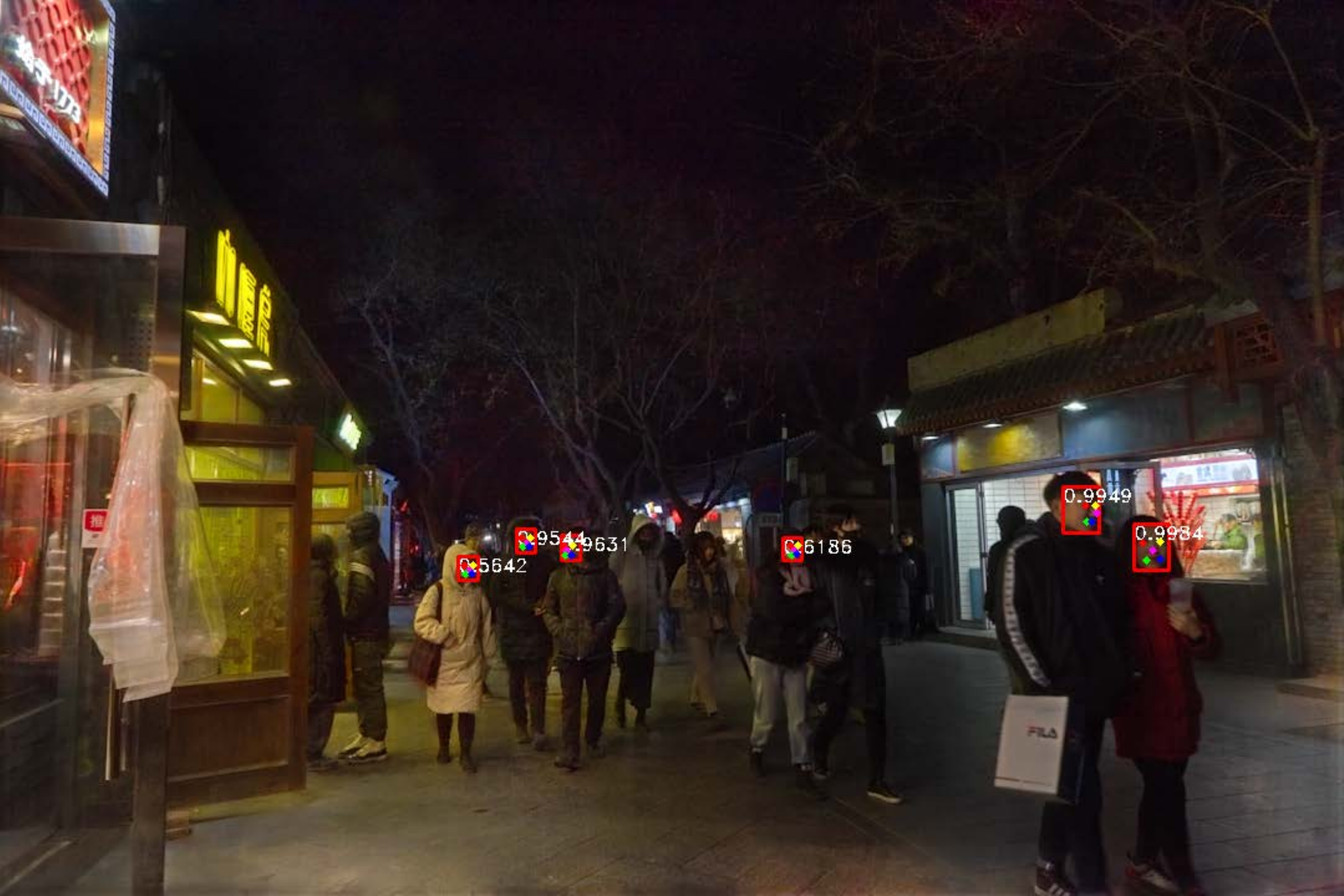}\\
   RetinaFace + input image & RetinaFace + Uformer \\
  \vspace{-1mm}
   \includegraphics[width=\swtwo]{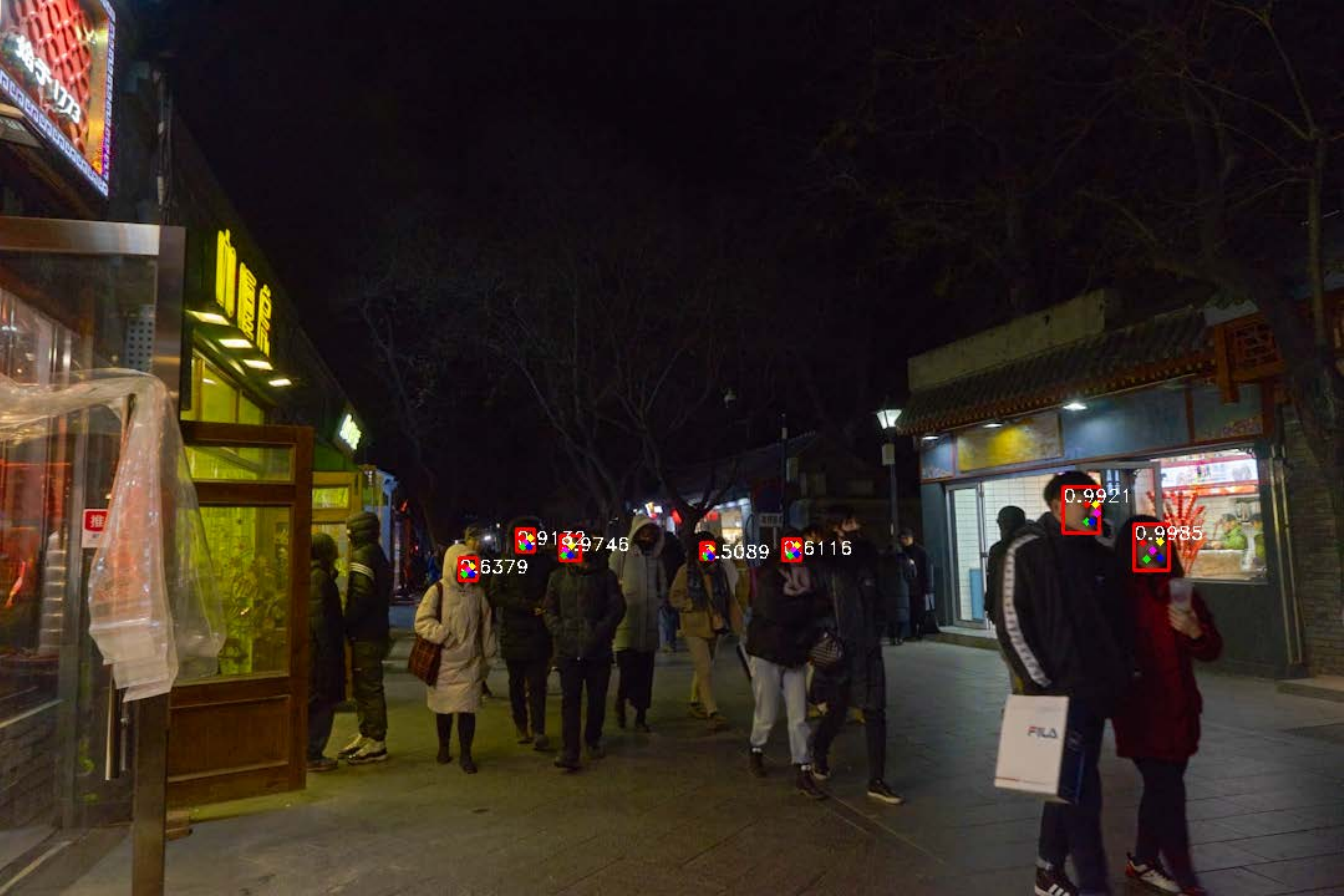}&
   \includegraphics[width=\swtwo]{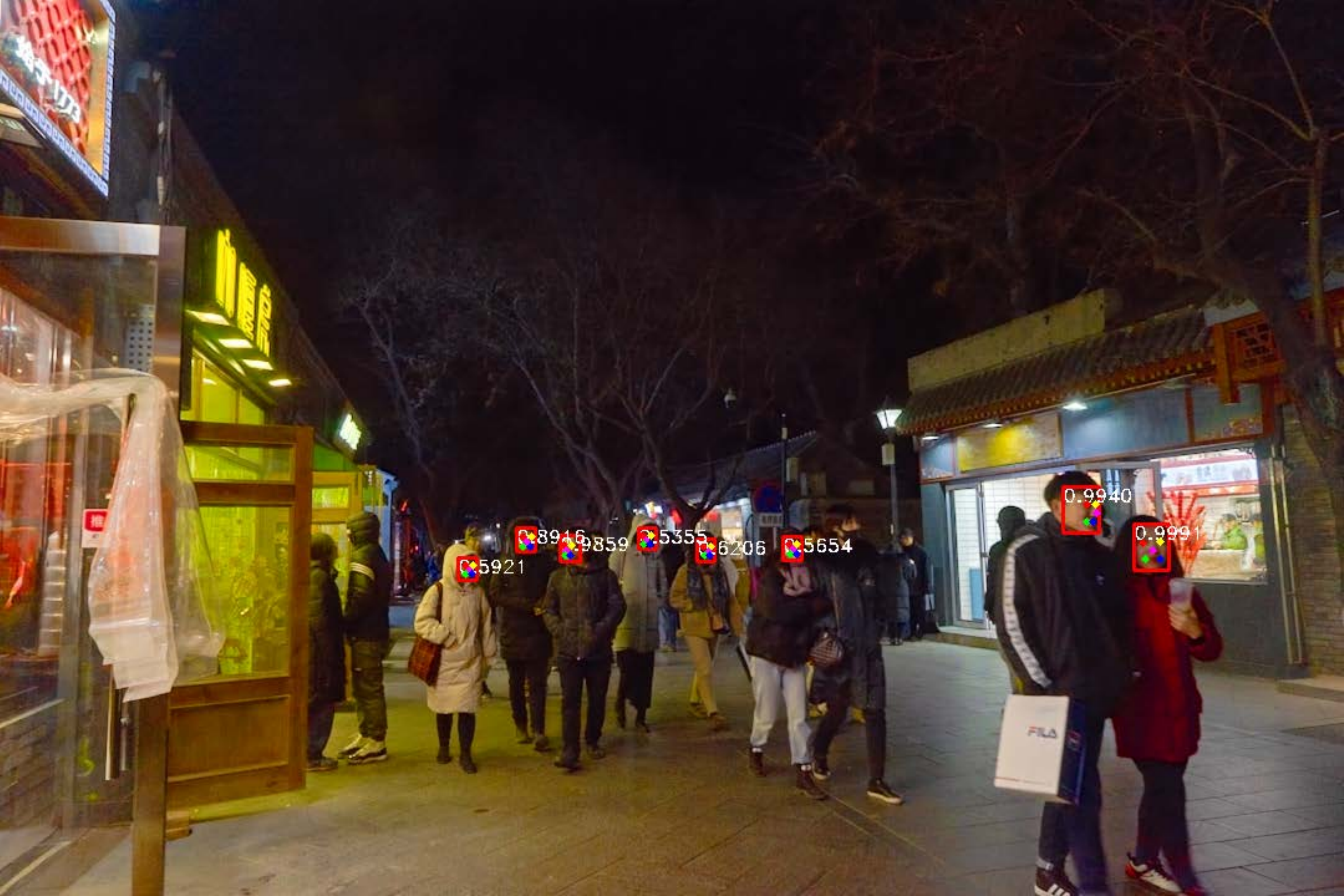}\\
    RetinaFace + Restormer &\textbf{RetinaFace + LLFormer}\\

\end{tabular}}
\end{center}
\vspace{-5mm}
\caption{Enhanced visual results and  face detection results.}
\label{fig:face_detection}
 \vspace{-0.2in}
\end{figure}%
\subsection{Benchmarking Study for UHD-LLIE}

\textbf{UHD-LOL4K Subset}. We test 16 different state-of-the-art LLIE methods and our proposed LLFormer on the UHD-LOL4K subset. 
The quantitative results are reported in Table \ref{tab:results1}. According to Table \ref{tab:results1}, we can find that traditional LLIE algorithms (BIMEF, FEA, LIME, MF, NPE, SRIE, MSRCR) generally do not work well on UHD-LOL4K. Among them, the quantitative scores (PSNR, SSIM, LPIPS, MAE) of some methods  are even worse than those of unsupervised learning methods (RUAS, ELGAN). The results of CNN-based supervised learning methods (see RetiunexNet, DSLR, and KID) are better than unsupervised learning-based and zero-shot learning-based methods, which is expected. Among the CNN-based methods, DSLR obtains the best performance in terms of PSNR, SSIM, LPIPS, and MAE. Compared with CNN-based supervised learning methods, the performances of transformer-based supervised learning methods (Uformer, Restormer, and LLFormer) are greatly improved. Among these, the proposed LLFormer obtains the best performance, achieving a $0.42$ dB improvement in PSNR compared to Restormer. A visual comparison is shown in Fig.~\ref{fig:cmp1vAll}. The image recovered by LLFormer contains vivid colors and is closer to the ground truth.

\textbf{UHD-LOL8K Subset}. We also conduct benchmarking experiments on the UHD-LOL8K subset by partitioning each 8K image into 4 patches of 4K resolution. The last four columns of Table \ref{tab:results1} show the evaluation results. Deep learning methods RetinexNet, DSLR, Uformer, Restormer, and LLFormer achieve better performance on both pixel-wise and perceptual metrics. Transformer-based methods achieve top ranks for all evaluation metrics with LLFormer outperforming other methods. As shown in Fig.~\ref{fig:cmp1vAll}, LLFormer produces visually pleasing results with more details.

\textbf{Improving Downstream Tasks}. To verify whether LLIE is beneficial for downstream tasks, we randomly select $300$ images from the DARK FACE dataset~\cite{yang2020advancing} and pre-process these images using the top three methods in our benchmark study.
 We then detect faces using RetinaFace~\cite{deng2020retinaface}. When using the pre-processing step, the average precision (AP) values for Uformer, Restormer, and LLFormer improve by 67.06\%, 68.11\%, and \textbf{71.2\%}, respectively. Visual results are shown in Fig.~\ref{fig:face_detection}. Pre-trained LLIE models not only generate images with adequate color balance, but also help improve the performance of downstream tasks.

\begin{table*}[t]\small
\begin{center}
\vspace{-0.1in}
\scalebox{0.7}{\begin{tabular}{l|c|c|c|c||c|c|c|c}
\hline
\multirow{2}{*}{Methods}  & \multicolumn{4}{c||}{\textbf{LOL}} & \multicolumn{4}{c}{\textbf{MIT-Adobe FiveK}}  \\ \cline{2-9} 

 & PSNR $\uparrow$ & SSIM $\uparrow$ & LPIPS $\downarrow$ & MAE $\downarrow$  & PSNR $\uparrow$ & SSIM $\uparrow$ & LPIPS $\downarrow$ & MAE $\downarrow$ \\ \hline

 BIMEF \cite{ying2017bio}  & 13.8752 & 0.5950 & 0.3264 & 0.2063  & 17.9683 & 0.7972 & 0.1398 & 0.1134 \\ \cline{2-9} 
 FEA \cite{dong2011fast}  & 16.7165 & 0.4784 & 0.3847 & 0.1421  & 15.2342 & 0.7161 & 0.1949 & 0.1512\\ \cline{2-9} 
LIME \cite{guo2016lime} & 16.7586 & 0.4449 & 0.3945 & 0.1200 &  13.3031 & 0.7497 & 0.1319 & 0.2044   \\ \cline{2-9} 
MF \cite{fu2016fusion}  & 16.9662 & 0.5075 &  0.3796  & 0.1416 &  17.6271& 0.8143 &  0.1204  & 0.1194 \\ \cline{2-9} 
 NPE \cite{wang2013naturalness} & 16.9697 &  0.4839  &  0.4049 & 0.1290&  17.3840 & 0.7932 & 0.1320 &  0.1224  \\ \cline{2-9} 
                    
SRIE \cite{fu2016weighted}  & 11.8552 & 0.4954 & 0.3401 & 0.2571 & 18.6273 & 0.8384 & 0.1047 & 0.1030   \\  \cline{2-9}
                                    
MSRCR \cite{jobson1997multiscale} &  13.1728& 0.4615 & 0.4350 & 0.2067 & 13.3149 & 0.7515 &0.1767& 0.1993 \\ \cline{2-9} 

RetinexNet \cite{wei2018deep}   & 16.7740 & 0.4250 & 0.4739& 0.1256 & 12.5146 &0.6708 &0.2535&0.2068   \\ \cline{2-9} 
                    
DSLR \cite{lim2020dslr}   & 14.9822 & 0.5964  & 0.3757&0.1918 & 20.2435 & 0.8289  & 0.1526 & 0.0880     \\ \cline{2-9} 
                    
KinD \cite{zhang2019kindling}   & 17.6476 & \textcolor{purple}{0.7715} & \textcolor{purple}{0.1750} & 0.1231 & 16.2032 & 0.7841& 0.1498&0.1379 \\ \cline{2-9}

Z\_DCE \cite{guo2020zero} &  14.8607 & 0.5624 & 0.3352 &0.1846 &  15.9312 & 0.7668  &0.1647& 0.1426 \\ \cline{2-9} 
                     
Z\_DCE++ \cite{Zero-DCE++} &  14.7484 & 0.5176  & 0.3284  & 0.1801& 14.6111 & 0.4055  & 0.2309 & 0.1539     \\ \cline{2-9}
                     
RUAS \cite{liu2021retinex} &   16.4047  &  0.5034 & 0.2701  &  0.1534 &  15.9953 & 0.7863 & 0.1397 & 0.1426     \\ \cline{2-9} 
                     
ELGAN \cite{jiang2021enlightengan} & 17.4829& 0.6515  & 0.3223  & 0.1352&  17.9050 &  0.8361 & 0.1425 & 0.1299      \\ \cline{2-9}
                     
Uformer \cite{wang2021uformer}  & \textcolor{purple}{18.5470} & 0.7212 &0.3205& 0.1134& \textcolor{purple}{21.9171}  &  \textcolor{purple}{0.8705} &  \textcolor{purple}{0.0854}  &  \textcolor{purple}{0.0702}
     \\ \cline{2-9} 
Restormer \cite{zamir2021restormer} & \textcolor{blue}{22.3652} & \textcolor{blue}{0.8157} & \textcolor{red}{0.1413}& \textcolor{blue}{0.0721} &  \textcolor{blue}{24.9228}
 &  \textcolor{blue}{0.9112} &   \textcolor{blue}{0.0579} &   \textcolor{blue}{0.0556}
    \\ \cline{2-9} 
                   
\textbf{LLFormer} &  \textcolor{red}{23.6491} & \textcolor{red}{0.8163} &  \textcolor{blue}{0.1692}& \textcolor{red}{0.0635}
  &\textcolor{red}{25.7528} & \textcolor{red}{0.9231} & \textcolor{red}{0.0447} & \textcolor{red}{0.0505}
     \\ \hline
\end{tabular}}
\vspace{-0.1in}
\caption{Comparison results on LOL and MIT-Adobe FiveK datasets in terms of PSNR, SSIM, LPIPS and MAE. The top three results are marked in \textcolor{red}{red}, \textcolor{blue}{blue} and \textcolor{purple}{purple}, respectively. Same as~\cite{zamir2020learning}, we consider images from expert C for the MIT-Adobe FiveK dataset.}
\label{tab:results2}
\vspace{-0.15in}
\end{center}
\end{table*}

\renewcommand{\tabcolsep}{.5pt}
\begin{figure*}[ht]
\vspace{-0.2cm}
\begin{center}
\begin{tabular}{ccccccccc}
   \vspace{-1.0mm}
   \includegraphics[width=\swnine]{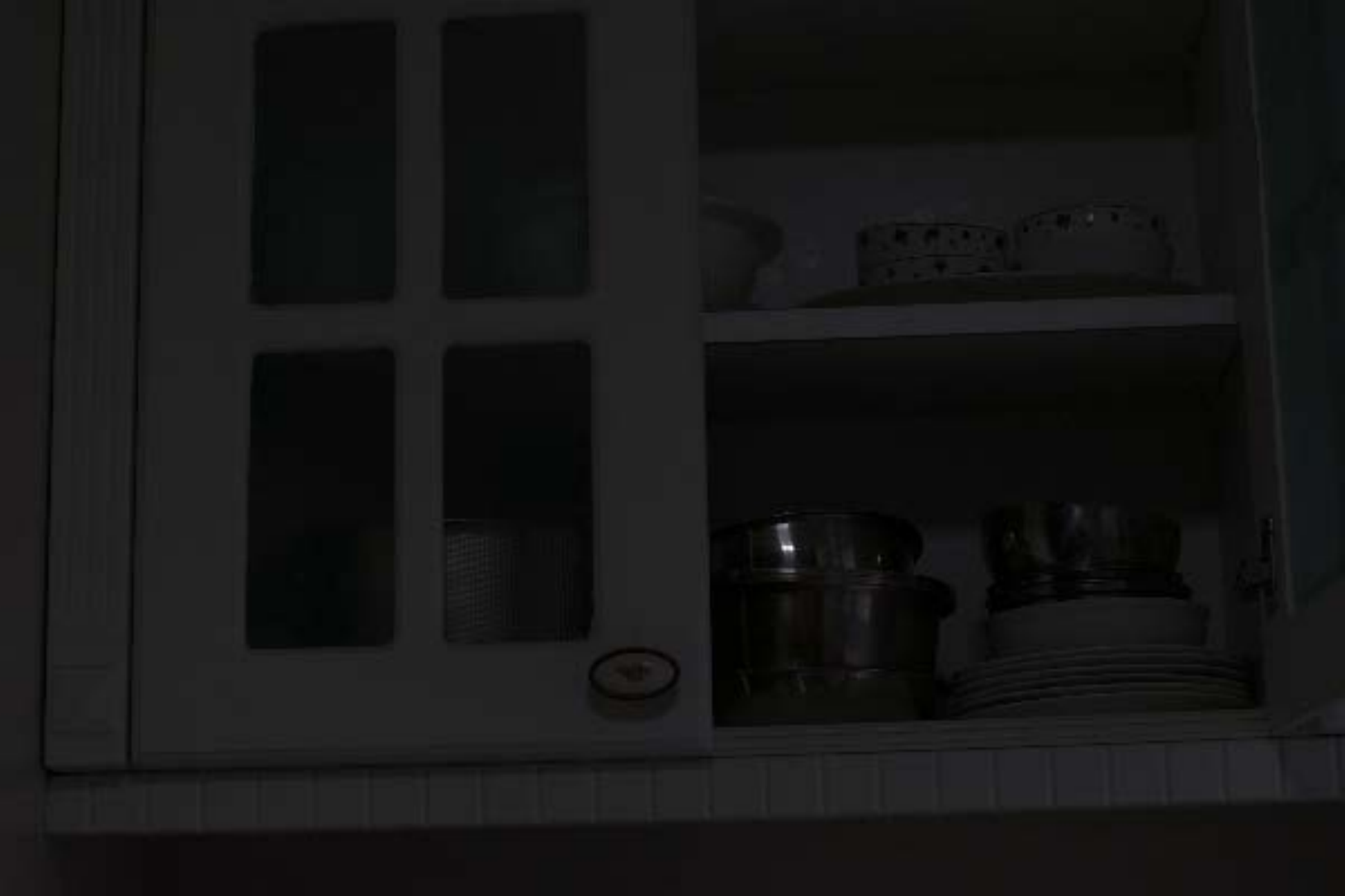}&
   \includegraphics[width=\swnine]{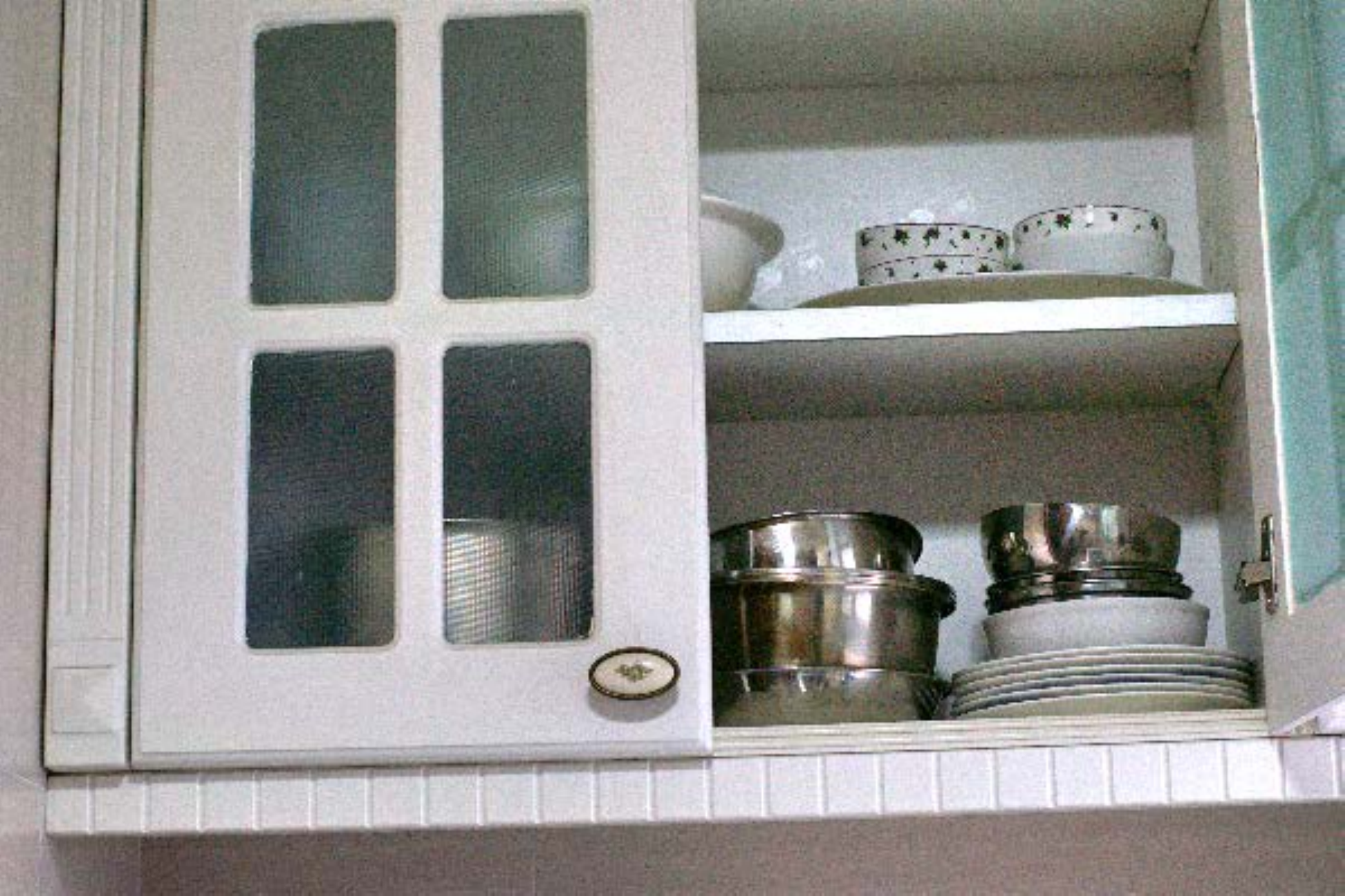}&
   \includegraphics[width=\swnine]{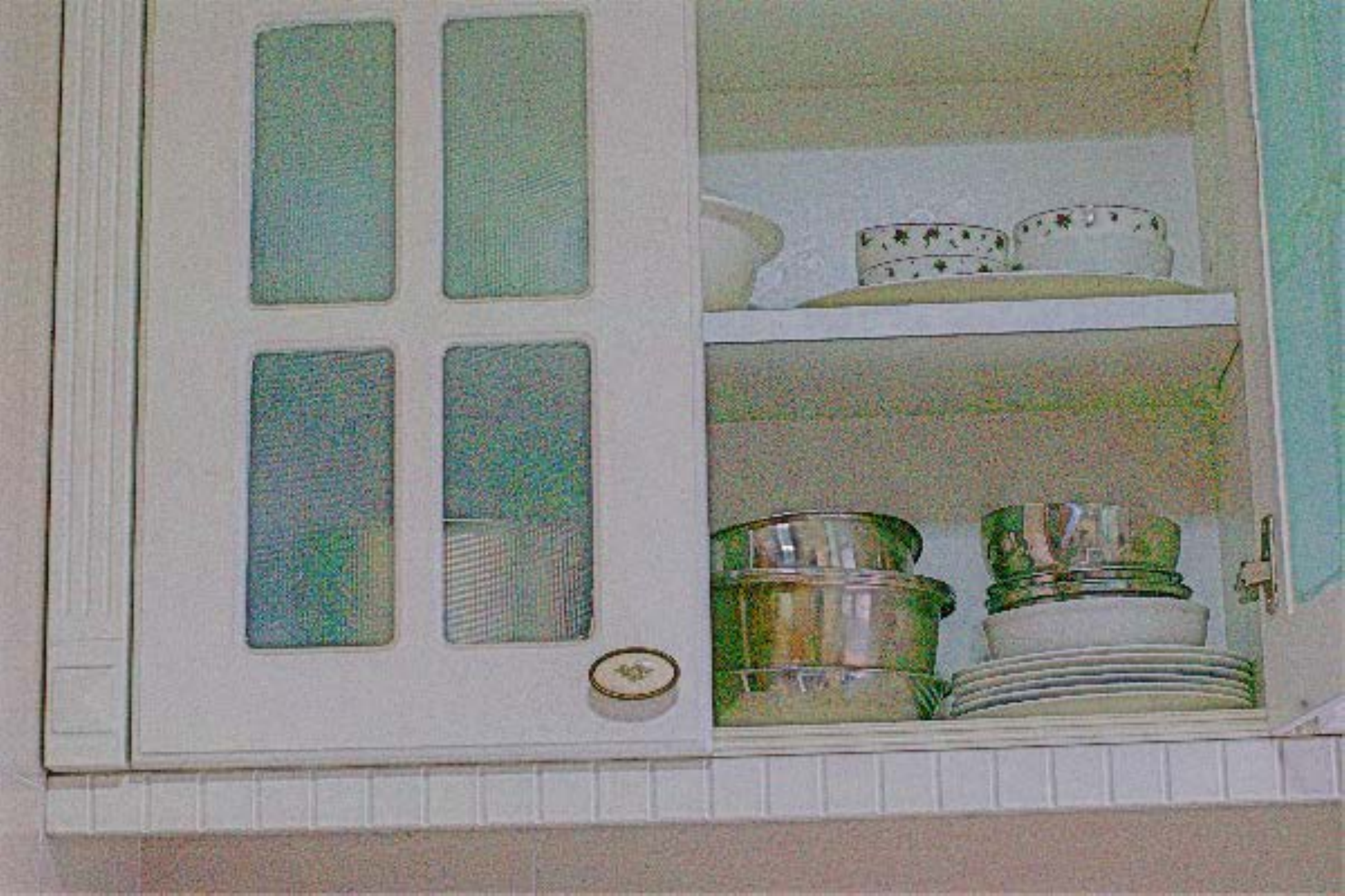}&
   \includegraphics[width=\swnine]{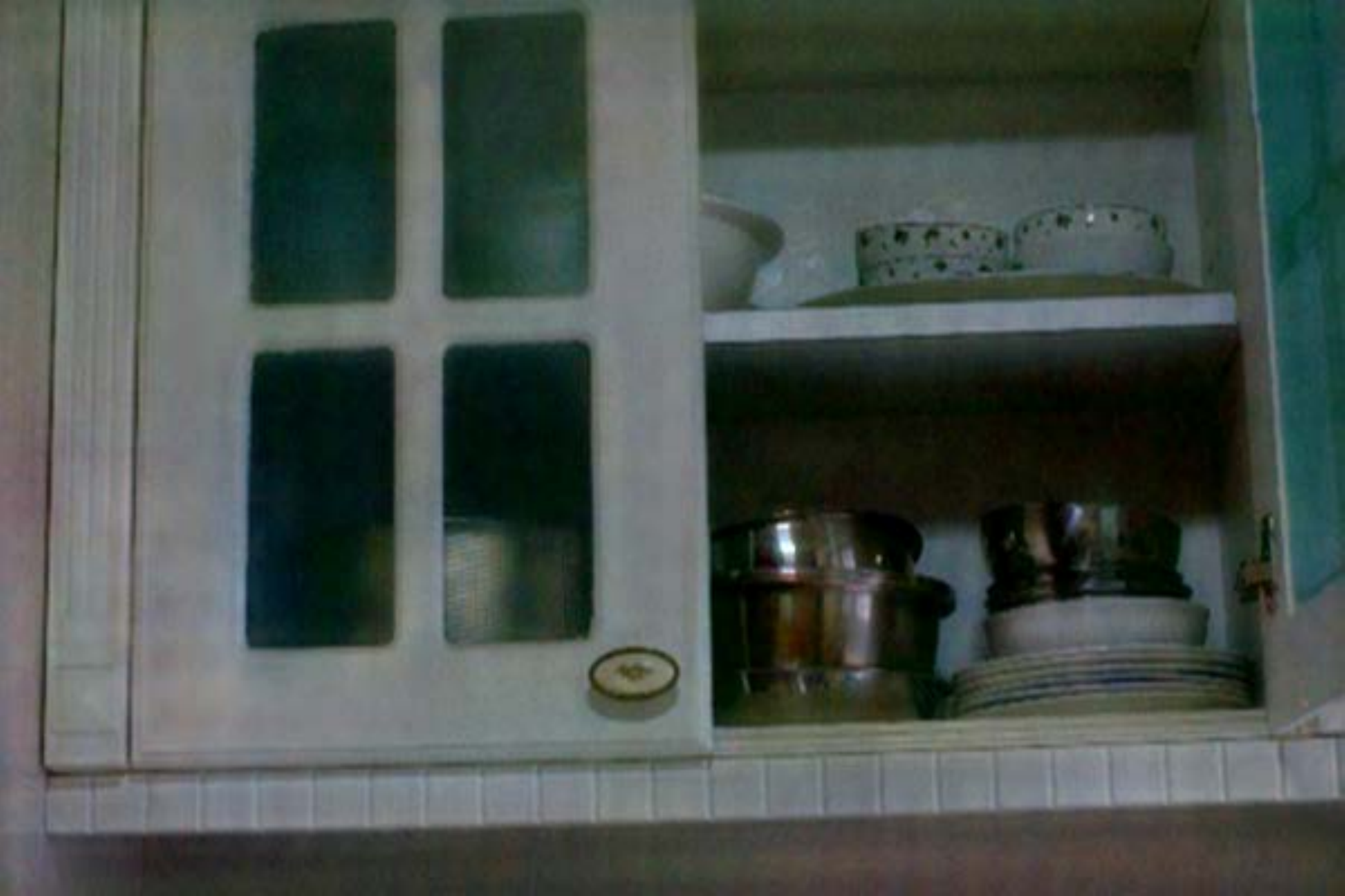}&
   \includegraphics[width=\swnine]{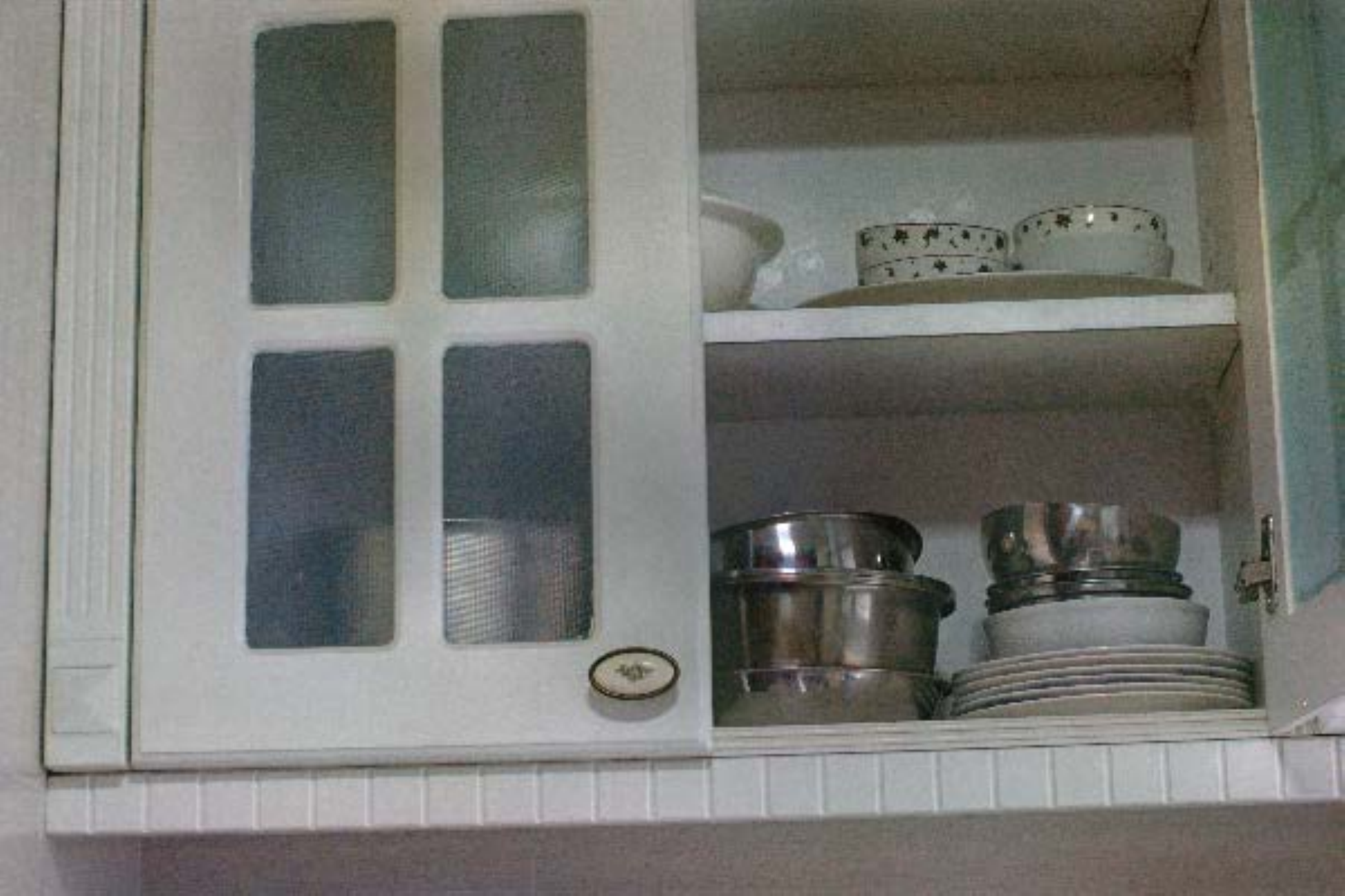}&
   \includegraphics[width=\swnine]{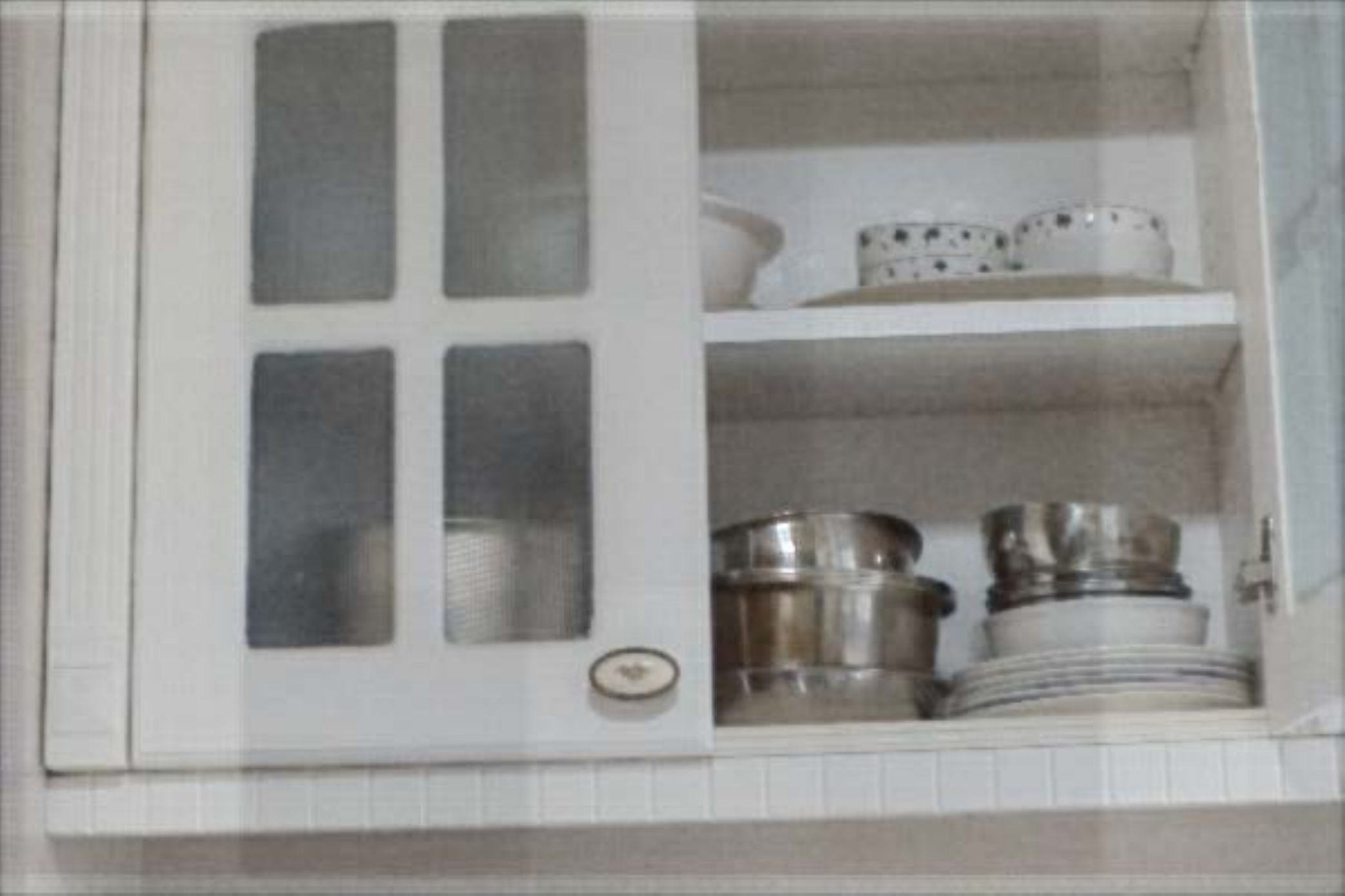}&
   \includegraphics[width=\swnine]{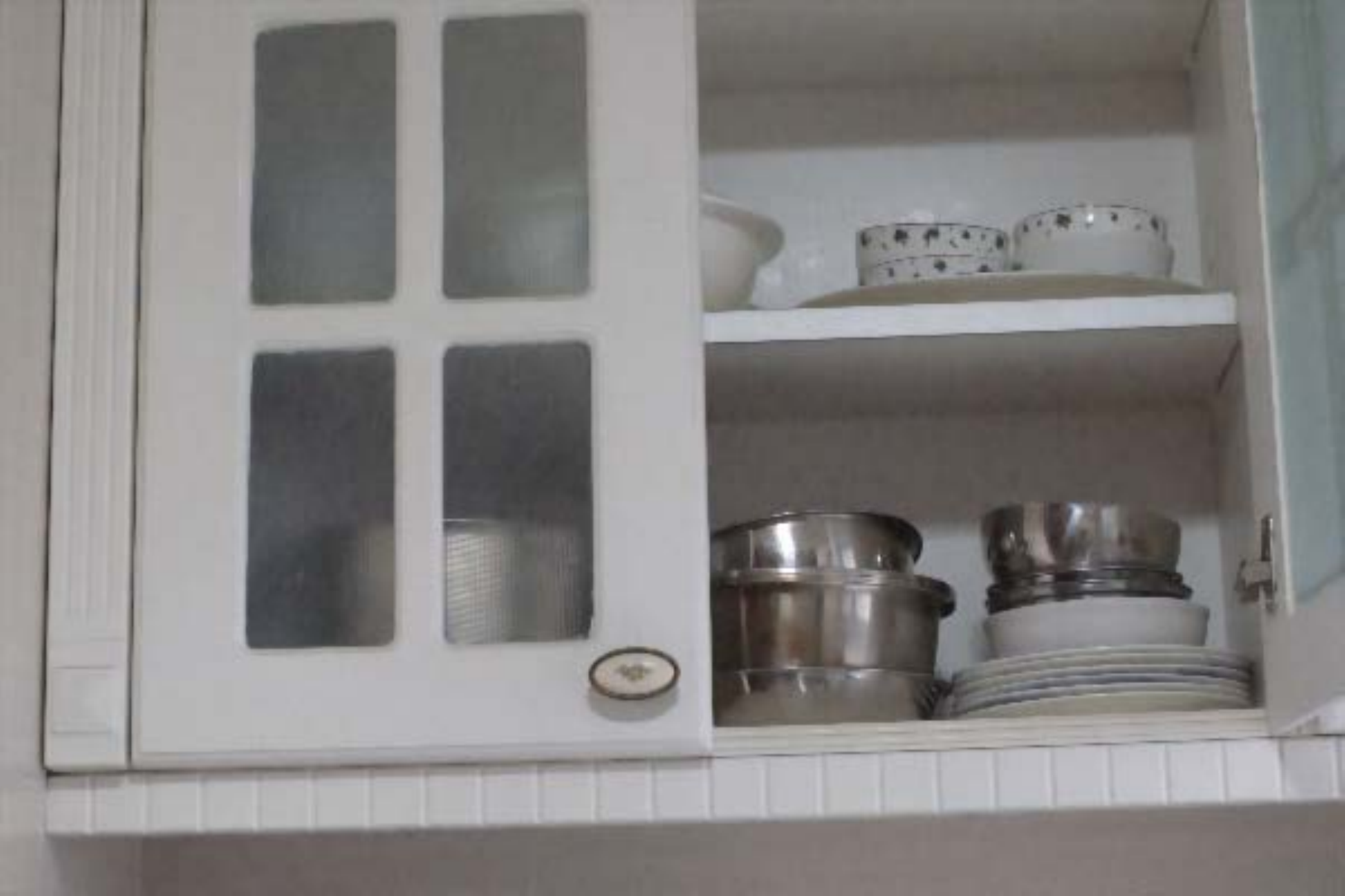} &
   \includegraphics[width=\swnine]{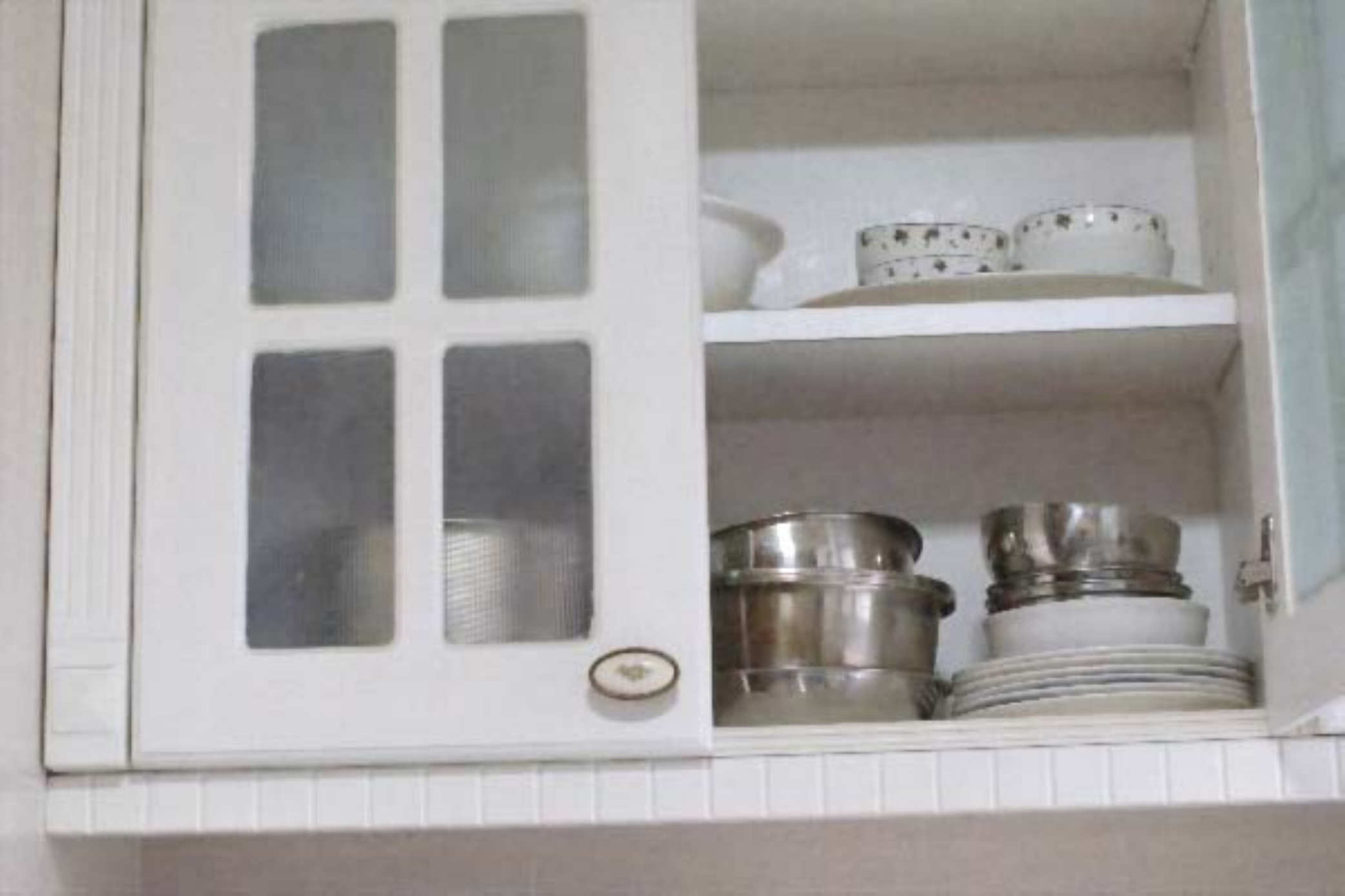} &
   \includegraphics[width=\swnine]{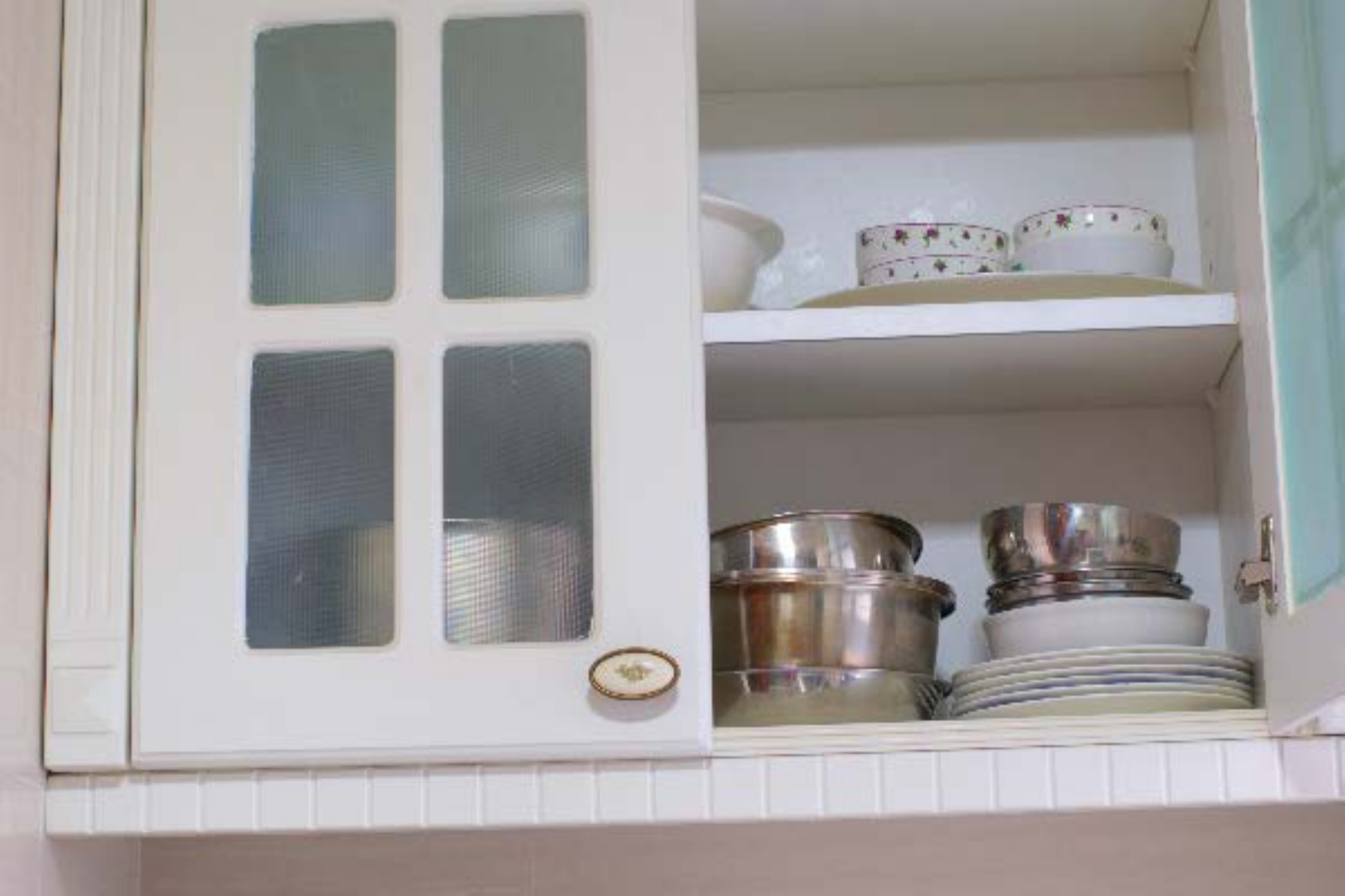} \\
   \vspace{-1.0mm}
   \includegraphics[width=\swnine]{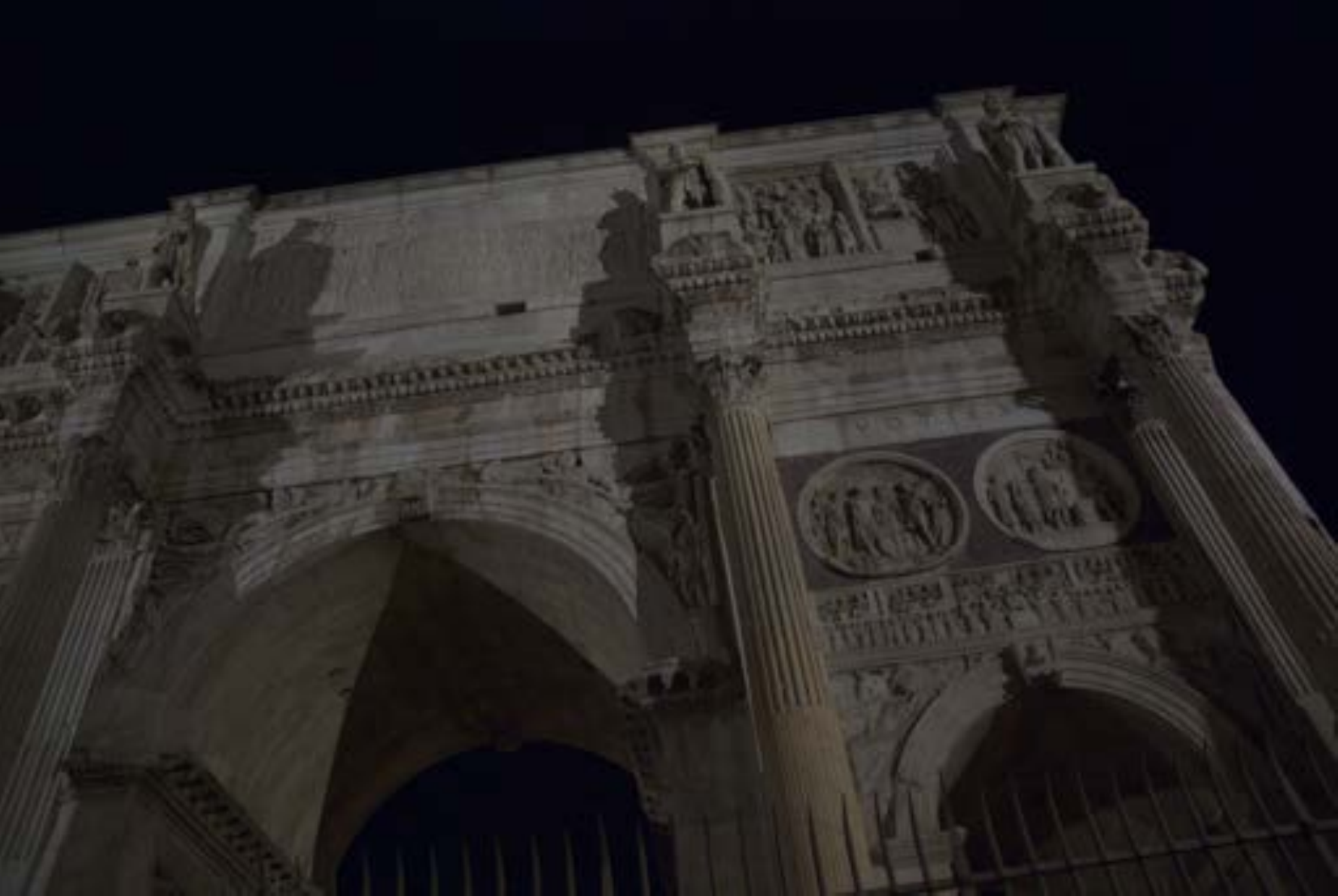}&
   \includegraphics[width=\swnine]{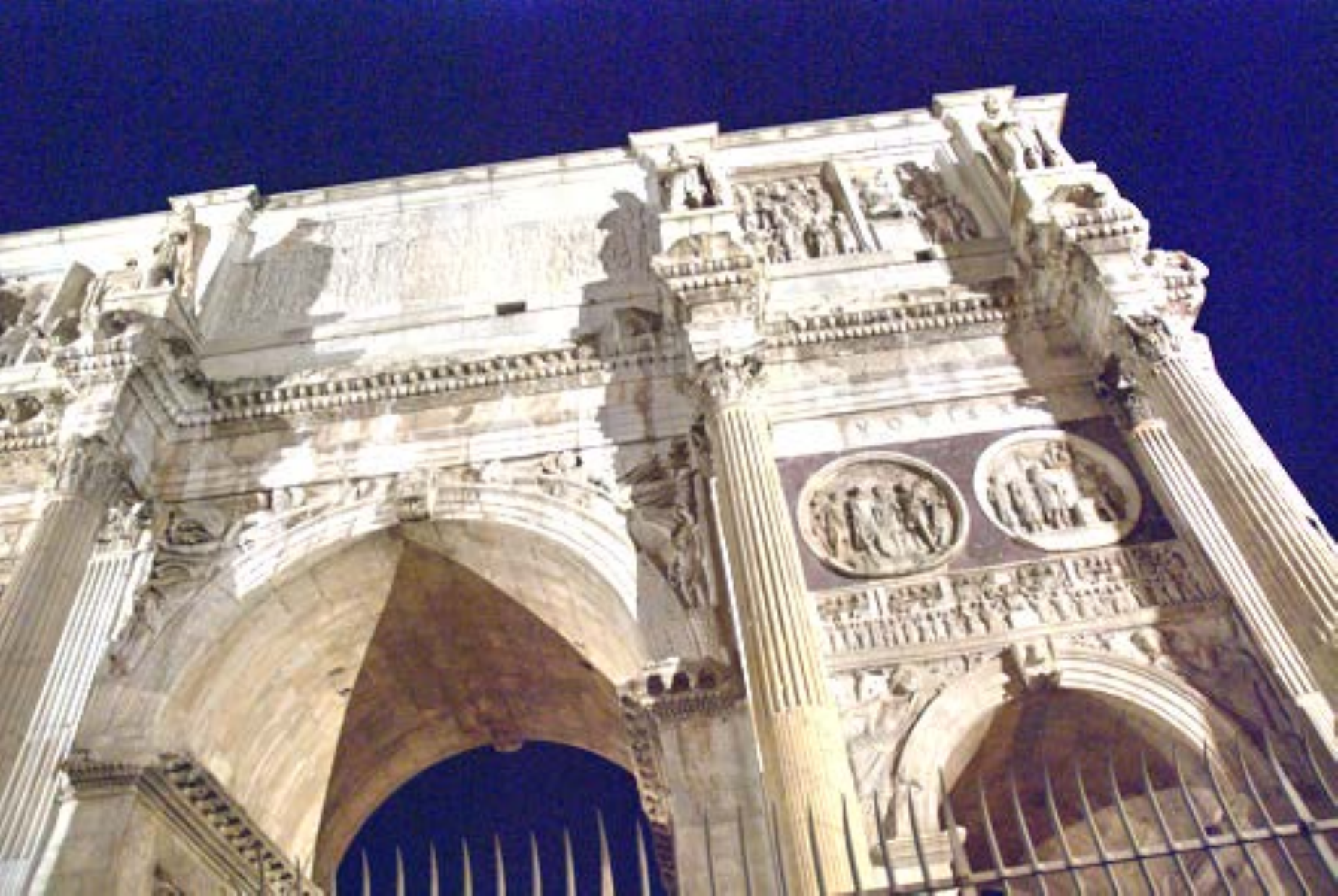}&
   \includegraphics[width=\swnine]{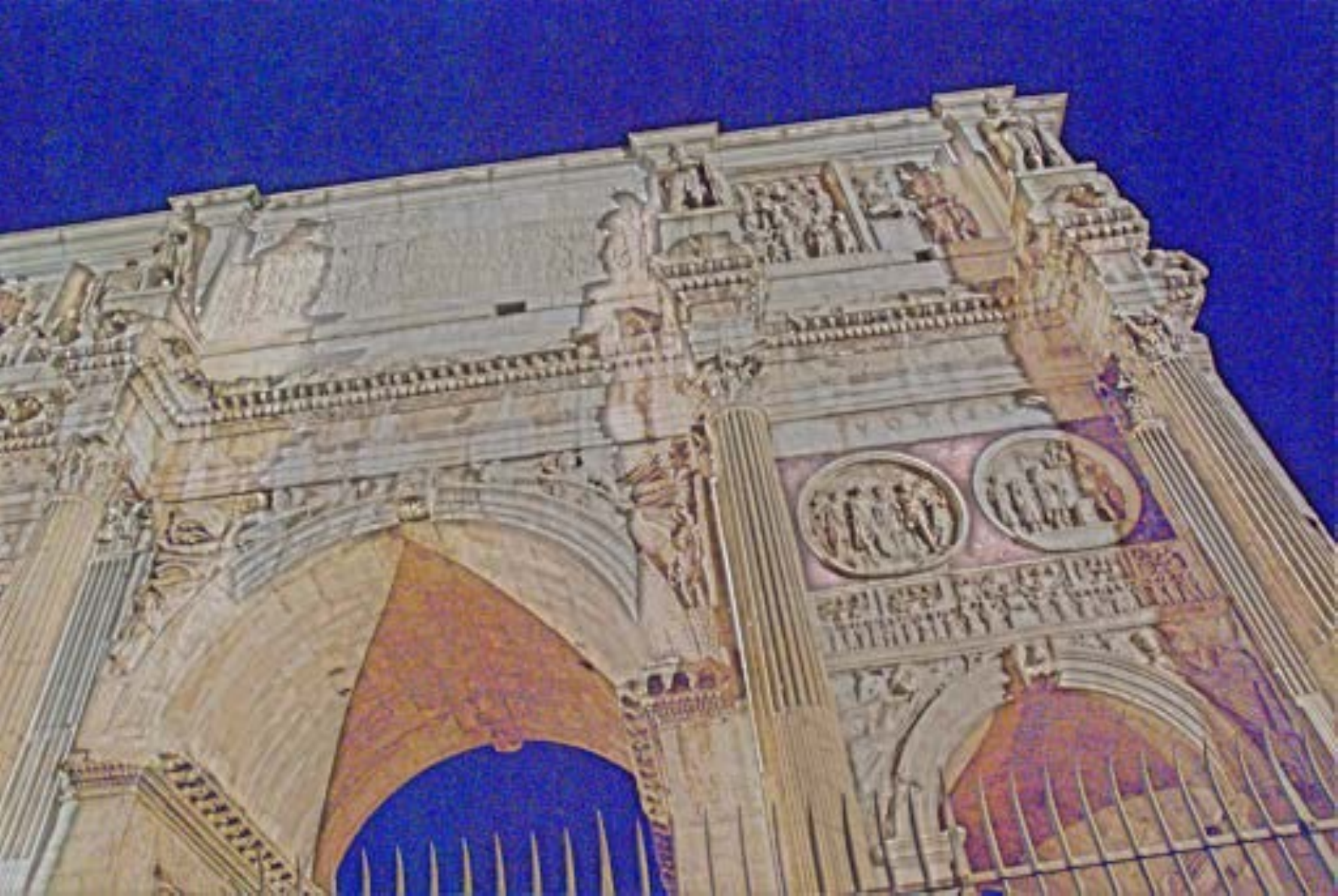}&
   \includegraphics[width=\swnine]{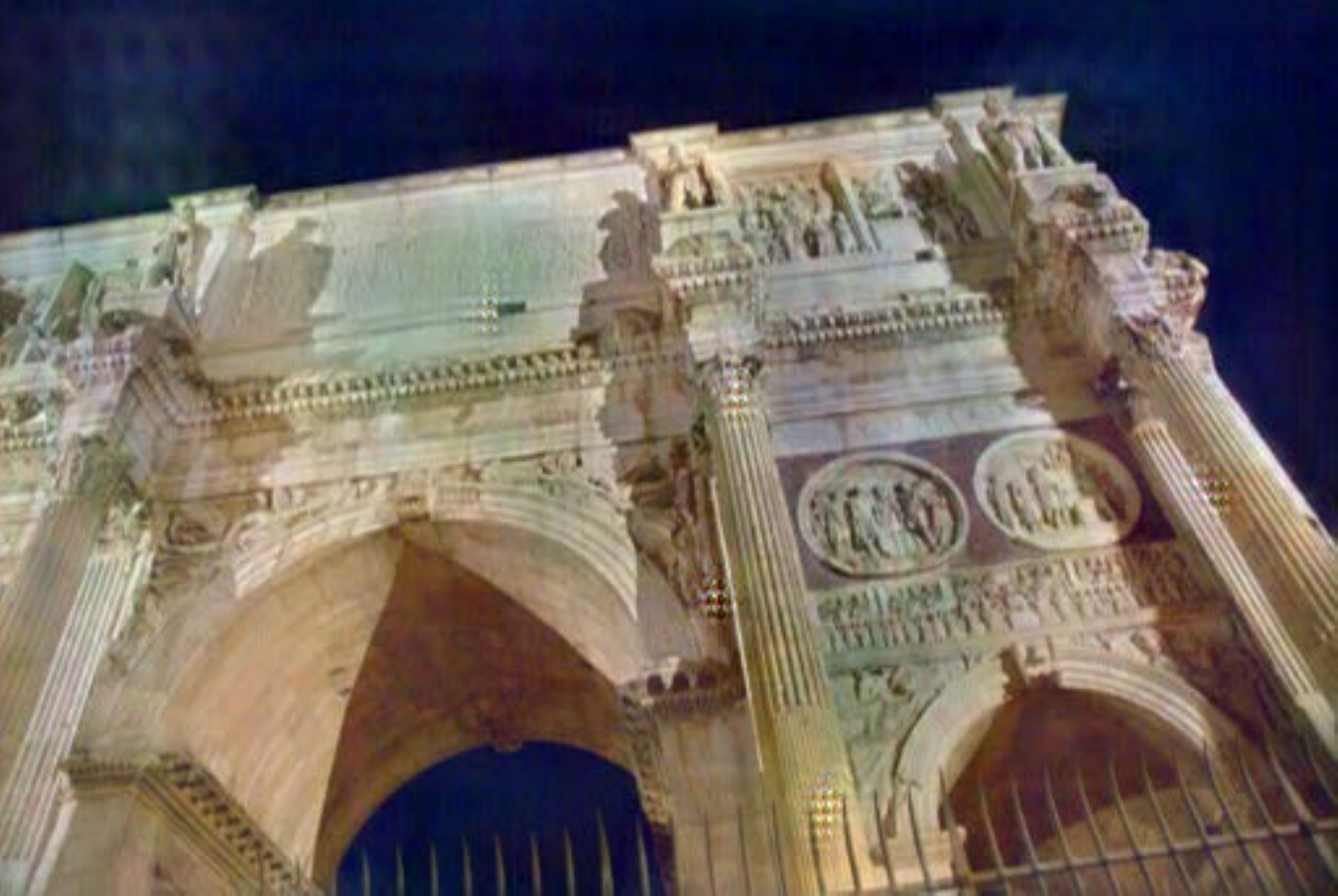}&
   \includegraphics[width=\swnine]{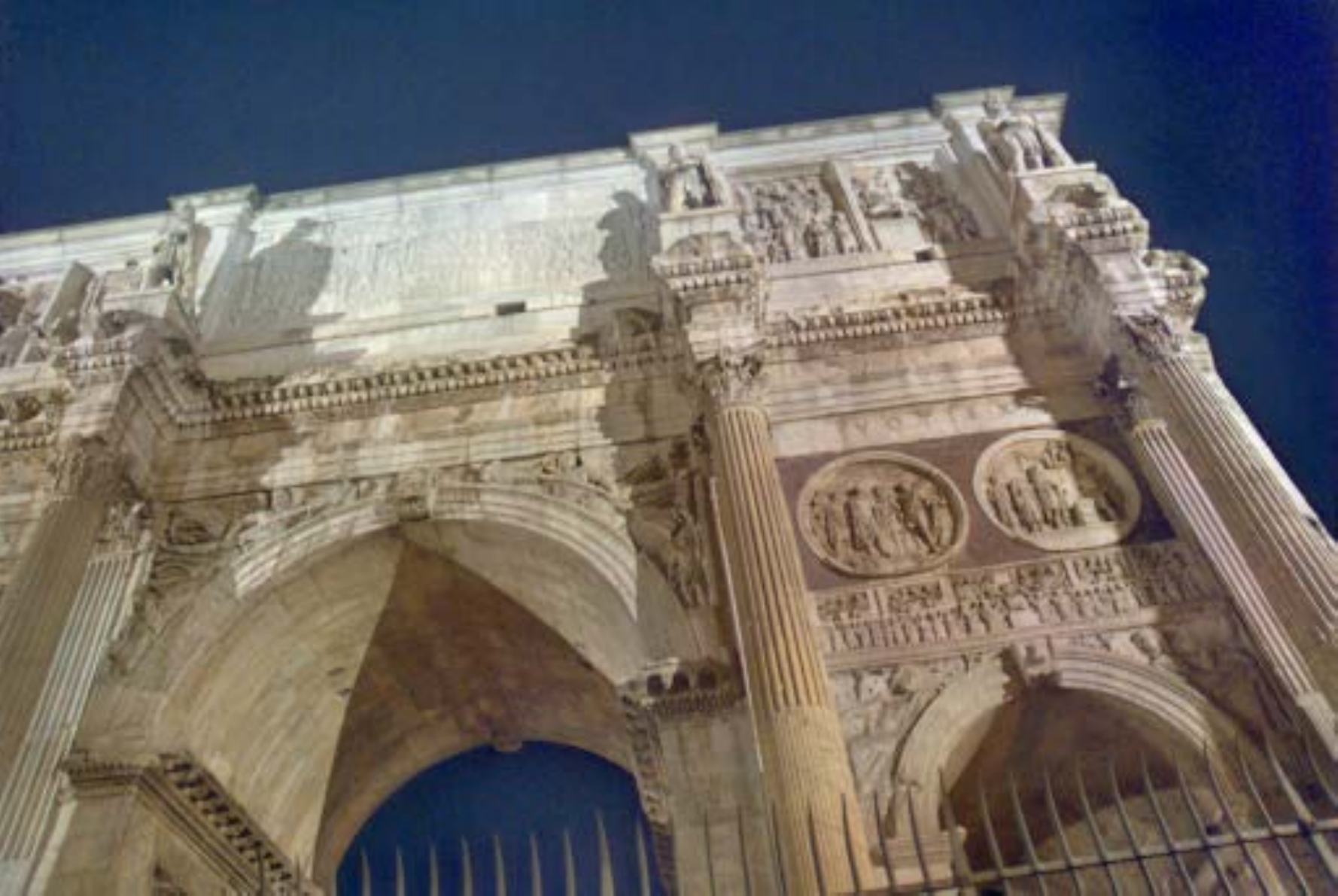}&
   \includegraphics[width=\swnine]{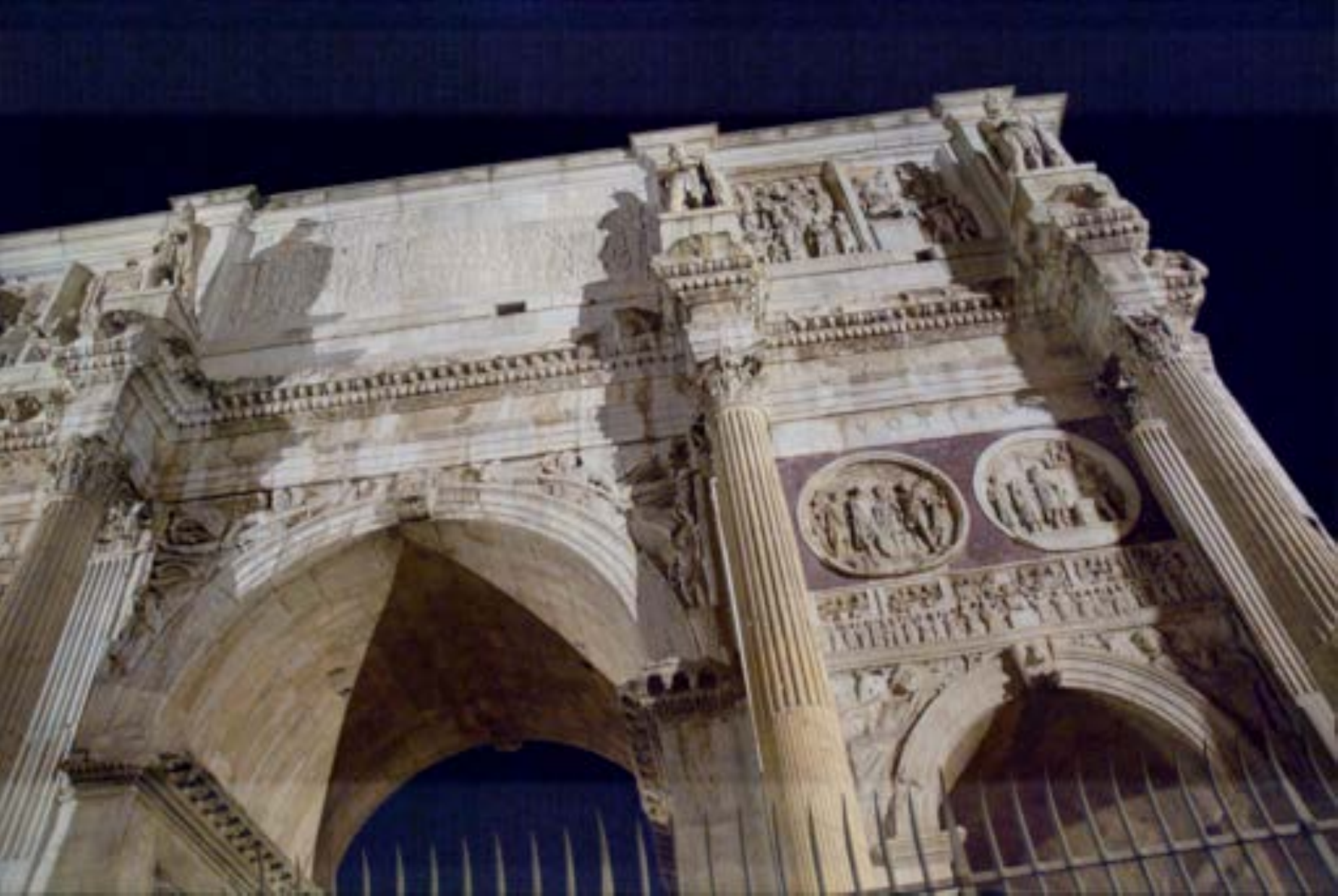}&
   \includegraphics[width=\swnine]{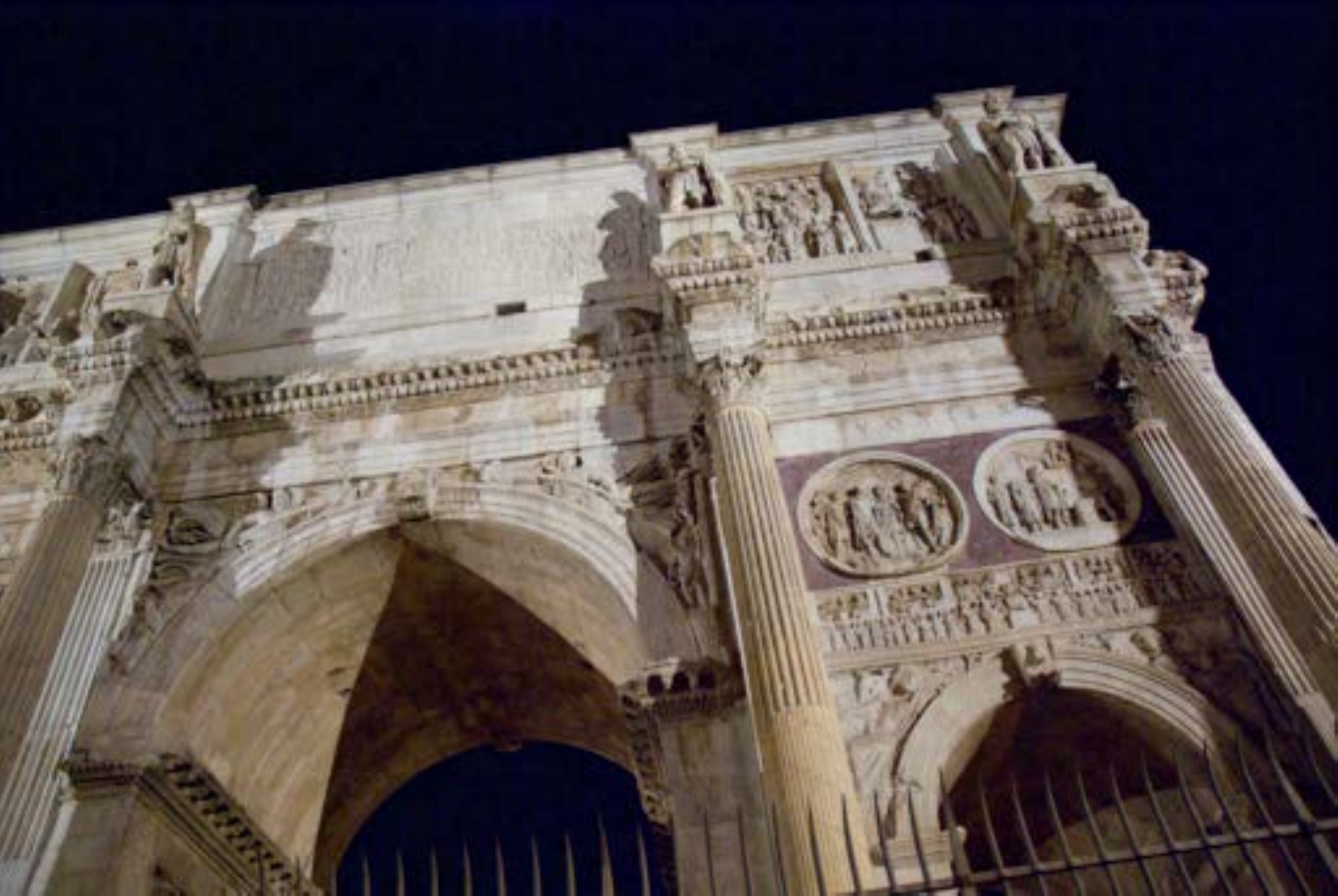} &
   \includegraphics[width=\swnine]{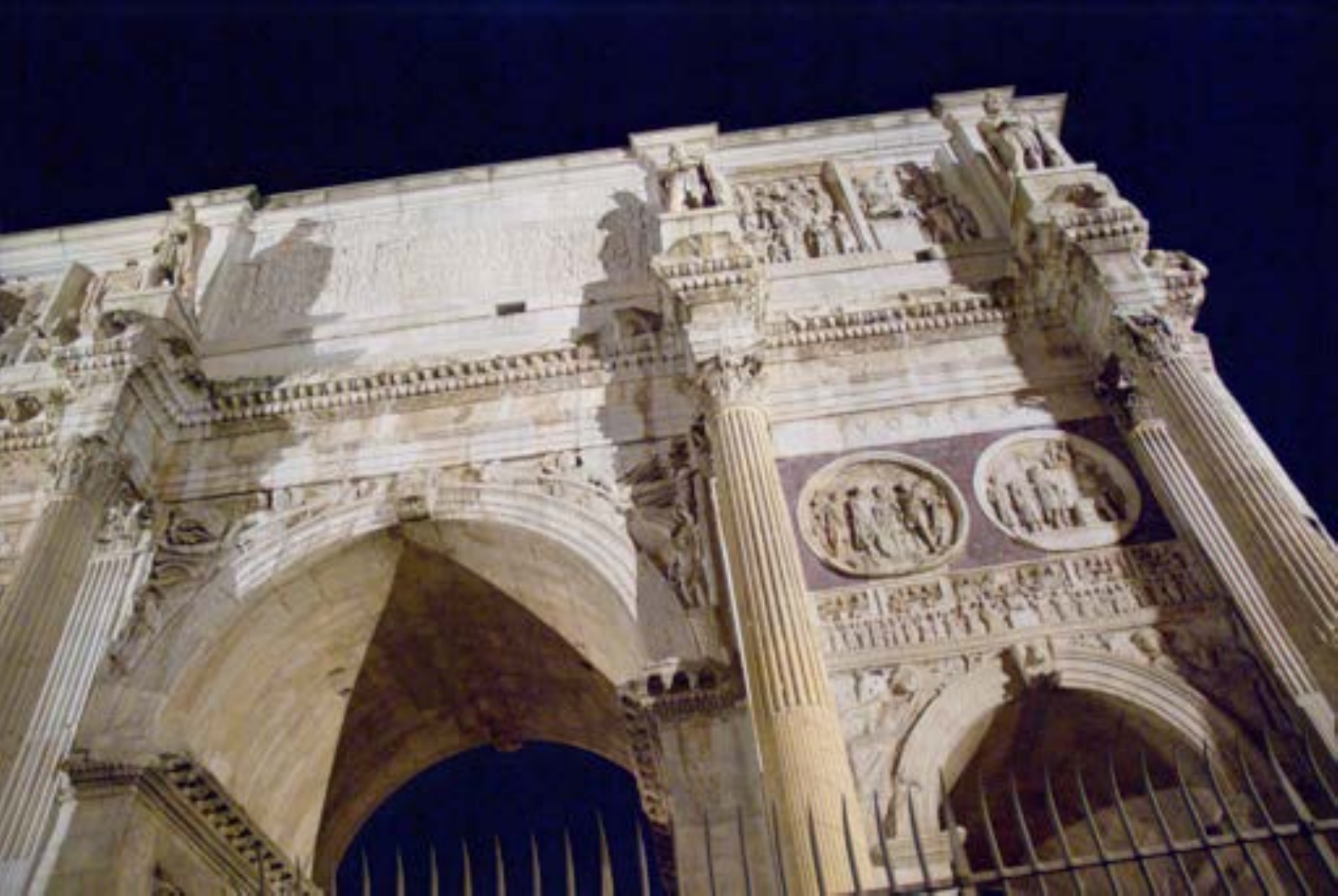}&
   \includegraphics[width=\swnine]{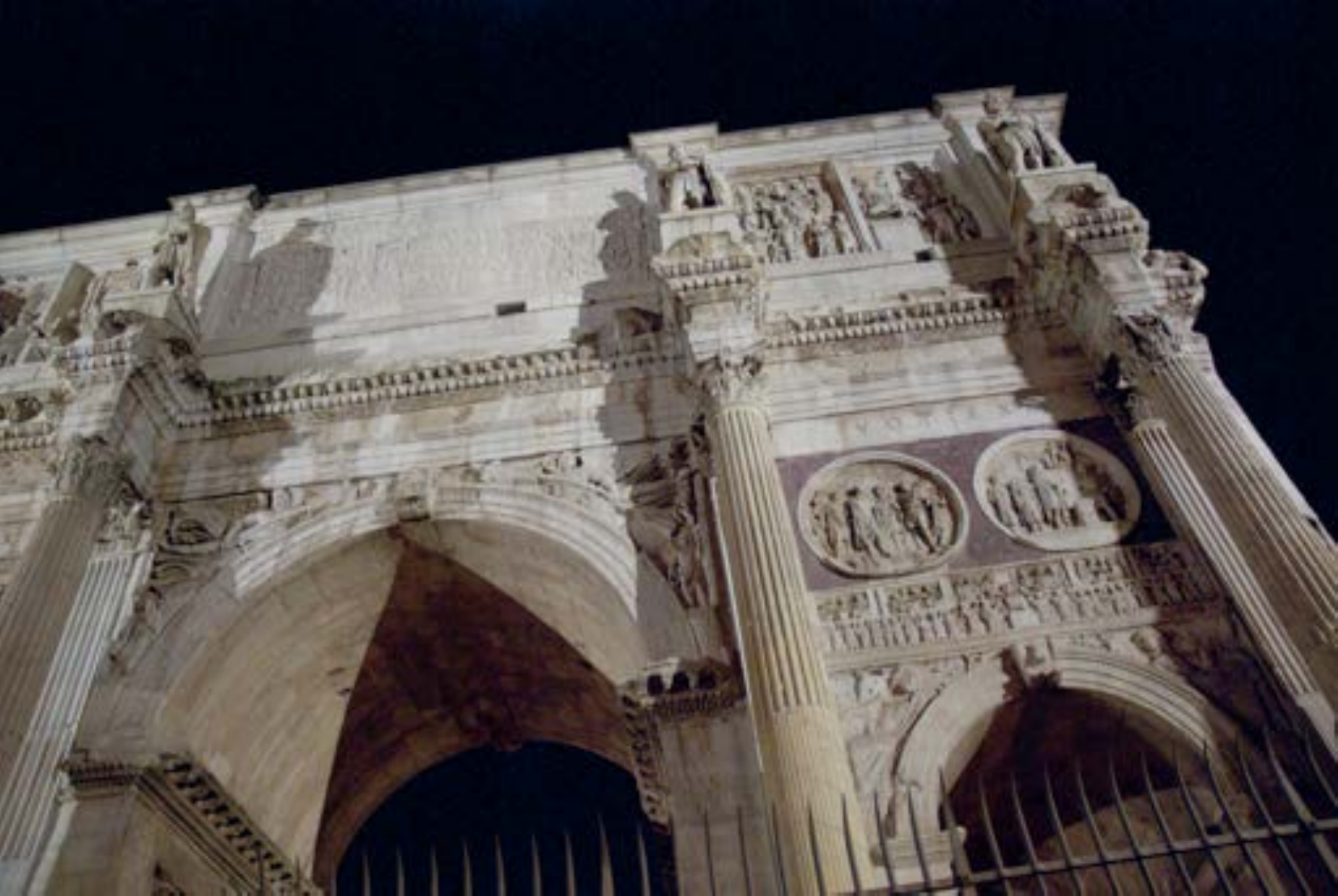} \\
   Input & LIME & RetinexNet  &  DSLR & ELGAN & Uformer & Restormer & \textbf{LLFormer} & GT
\end{tabular}
\end{center}
\vspace{-5mm}
\caption{Visual comparison. The top row is from the LOL dataset and the bottom is from the MIT-Adobe FiveK dataset. \textbf{Zoom in for a better view}.}
\label{fig:cmp1vAll1}
\vspace{-0.2in}
\end{figure*}%

\begin{table}[ht]\small
  \centering
    \scalebox{0.62}{\begin{tabular}{c|cc|c||c|c|c|c}
    \hline
   \textbf{Variant} & \multicolumn{4}{c|}{\textbf{Component}} & \textbf{MACs} & \textbf{Params} & \multicolumn{1}{c}{\textbf{PSNR/SSIM}} \\ \hline
    Base & \hypertarget{abla1}{\textcolor{blue}{(a)}} &\multicolumn{3}{c|}{ model with Resblock \cite{lim2017enhanced}
					} & 11.90G & 13.87M &31.92/0.9771  \\  \hline
   
 \multirow{4}{*}{\makecell[c]{Multi-head \\attention }} &\hypertarget{abla2}{\textcolor{blue}{(b)}} &\multicolumn{3}{c|}{A-MSA (Height) + DGFN} & 13.60G & 14.77M & 35.15/0.9836 \\ \cline{2-8}
  
& \hypertarget{abla3}{\textcolor{blue}{(c)}}& \multicolumn{3}{c|}{A-MSA (Width) + DGFN} & 13.60G & 14.77M & 34.98/0.9832 \\  \cline{2-8}    

& \hypertarget{abla4}{\textcolor{blue}{(d)}}& \multicolumn{3}{c|}{A-MSA + DGFN} & 16.26G & 19.78M & 35.31/0.9843 \\ \cline{2-8}

&\hypertarget{abla5}{\textcolor{blue}{(e)}} & \multicolumn{3}{c|}{A-MSA + FFN \cite{vaswani2017attention}} & 18.90G & 20.62M & 23.33/0.9111 \\ \hline

\multirow{2}{*}{\makecell[c]{Feed-Forward \\Network}} &\hypertarget{abla6}{\textcolor{blue}{(f)}} & \multicolumn{3}{c|}{A-MSA + DGFN} & 21.47G & 24.18M & 35.83/0.9846 \\
\cline{2-8}  

& \hypertarget{abla7}{\textcolor{blue}{(g)}}& \multicolumn{3}{c|}{ A-MSA + DGFN} & 25.52G & 24.52M & 32.78/0.9786 \\ \hline
 \rowcolor{Gray} LLFormer & \hypertarget{abla8}{\textcolor{blue}{(h)}}&\multicolumn{3}{c|}{A-MSA + DGFN} & \cellcolor{Gray} 22.52G & 24.52M & \textbf{36.20/0.9867} \\ \hline

    \end{tabular}}
\vspace{-0.1in}
 \caption{Ablation study on Transformer Block. (d) refers to A-MSA without depth-wise convolution, (f) is DGFN without depth-wise convolution, and (g) is DGFN without the dual gated mechanism.}
  \label{tab:addlabel}
    \vspace{-0.1in}
\end{table}%

\begin{table}[ht]\small
\begin{center}
    \scalebox{0.62}{\begin{tabular}{|c|c|c|c|c|c|}
\hline
   \textbf{Variant}  & \multicolumn{2}{c|}{\textbf{Component}} & \multicolumn{1}{c|}{\textbf{MACs}}& \multicolumn{1}{c|}{\textbf{Params}} & \multicolumn{1}{c|}{\textbf{PSNR/SSIM}}\\ \hline

 \multirow{3}{*}{Skip}  & \multicolumn{2}{c|}{w/o skip} & 22.52G & 24.52M &35.45/0.9844 \\  \cline{2-6}    

& \multicolumn{2}{c|}{w/o conv } & 22.47G & 24.50M & 35.90/0.9852\\ \cline{2-6}

 & \multicolumn{2}{c|}{w/o skip+conv} & 22.47G& 24.50M& 35.12/0.9847\\ \hline

\multirow{3}{*}{CAFB} & \multicolumn{2}{c|}{w/o head CAFB} & 21.76G & 24.51M & 35.57/0.9847\\
\cline{2-6}  

 & \multicolumn{2}{c|}{w/o tail CAFB} &  21.76G & 24.51M & 35.81/0.9852\\ \cline{2-6}  
 & \multicolumn{2}{c|}{w/o CAFB} &  21.00G & 24.50M & 35.10/0.9835\\ \hline
 \rowcolor{Gray} LLFormer  & \multicolumn{2}{c|}{contain all} & 22.52G & 24.52M& \textbf{36.20/0.9867} \\  
 \hline
    \end{tabular}}
    \vspace{-0.1in}
\caption{Ablation study on connection and fusion.}
\label{tab:component}
\vspace{-0.35in}
\end{center}
\end{table}

\begin{table}[ht]\small
 \begin{center}
   \scalebox{0.62}{\begin{tabular}{|c|c|c|c|c|c|c|}
   \hline
   \textbf{Variant} & \multicolumn{2}{c|}{\textbf{W/D}} & \textbf{MACs} & \textbf{Params} & \multicolumn{1}{c|}{\textbf{PSNR/SSIM}} & \multicolumn{1}{c|}{\textbf{Speed}} \\ \hline
\rowcolor{Gray} LLFormer & \multicolumn{2}{c|}{\textbf{16/4}} & 22.52G &  24.52M & \textbf{36.20/0.9867}  & 0.063 s \\  \hline
   
 \multirow{3}{*}{Wider} & \multicolumn{2}{c|}{\textbf{32}/4} & 81.92G & 95.63M &36.91/0.9871 &0.120 s \\  \cline{2-7}    

& \multicolumn{2}{c|}{\textbf{48}/4} & 111.22G & 114.49M & 37.44/0.9880&0.152 s \\ \cline{2-7}

& \multicolumn{2}{c|}{\textbf{64}/4} & 311.16G & 377.64M & 38.00/0.9881 & 0.193 s\\ \hline

\multirow{3}{*}{Deeper} & \multicolumn{2}{c|}{16/\textbf{3}} & 14.88G & 3.51M &20.19/0.9432 &0.054s \\
\cline{2-7}  
& \multicolumn{2}{c|}{16/\textbf{5}} & 29.53G & 106.77M &36.09/0.9844 &0.142 s \\ \cline{2-7}
 & \multicolumn{2}{c|}{ 16/\textbf{6}} & 36.45G & 432.25M &35.62/0.9847 & 0.181 s\\ \cline{2-7}  
 & \multicolumn{2}{c|}{ 16/\textbf{7}} & 43.32G& 1727.08M & 35.41/0.9845 & 0.217 s \\  
\hline
    \end{tabular}}
\vspace{-0.1in}
\caption{"Wider vs. Deeper" analysis.}
\label{tab:deep_wider}
\vspace{-0.35in}
\end{center}
\end{table}

\subsection{Comparison results on Public Datasets}
We benchmark LLFormer on the LOL \cite{wei2018deep} and MIT-Adobe FiveK \cite{bychkovsky2011learning} datasets, comparing it with  14 methods specifically designed for LLIE and two transformer-based  methods.
We use published code to retrain Uformer and Restormer on these datasets, respectively.  Results are shown in Table \ref{tab:results2}.
LLFormer achieves significantly higher performance on the LOL dataset, obtaining higher PSNR, SSIM, and MAE scores than Restormer.
In terms of LPIPS, LLFormer ranks in second place. 
On the MIT-Adobe FiveK dataset, transformer-based methods rank at the top and LLFormer achieves the best results on all metrics.
Among the best three transformer-based methods, the overhead (parameters and multiply-accumulate operations) for Uformer, Restormer and LLFormer are 38.82M/76.67G, 26.10M/140.99G and \textbf{24.52M}/\textbf{22.52G} (measured on $256 \times 256$ images), respectively. 
This shows that the proposed LLFormer achieves the best performance with efficient use of resources. This is due to the design of LLFormer, where the axis-based multi-head self-attention and hierarchical structure help to decrease the computational complexity. A visual comparison is shown in Fig.~\ref{fig:cmp1vAll1}. LLFormer produces images with adequate saturation as well as color and texture fidelity.

\subsection{Ablation Studies}
We conduct ablation studies by measuring the contributions of the following factors: (1) Axis-based Multi-head Self Attention; (2) Dual Gated Feed-Forward Network; (3) Weighted skip connection; (4) Cross-layer Attention Fusion Block; (5) Width and depth of the network. Experiments are performed on the UHD-LOL4K subset, and models are trained on image patches of size $128 \times 128$ for $100$ epochs.

\textbf{A. Axis-based Transformer Block}. 
We measure the impact of the proposed axis-based multi-head self attention and dual gated feed-forward network (FFN), see Table \ref{tab:addlabel}. 
Compared with the base model using Resblock~\cite{lim2017enhanced}, our A-MSA (either height or width) and DGFN significantly contribute to the improvements. 
When using depth-wise convolution to enhance locality in self-attention (compare \hyperlink{abla4}{\textcolor{blue}{(d)}} and ~\hyperlink{abla8}{\textcolor{blue}{(h)}}) or the feed-forward network (compare~\hyperlink{abla6}{\textcolor{blue}{(f)}} and \hyperlink{abla8}{\textcolor{blue}{(h)}}), the improvements in terms of PSNR are $0.89$, $0.75$, respectively. 
By applying the dual gated mechanism, PSNR and SSIM are improved by $3.42$ and $0.0081$ (see \hyperlink{abla6}{\textcolor{blue}{(g)}}~\hyperlink{abla8}{\textcolor{blue}{(h)}}).
Using the dual gated mechanism together with locality yields the best results. 
In contrast, combining A-MSA with the conventional FFN \cite{vaswani2017attention}, degrades the performance (Table \ref{tab:addlabel}~\hyperlink{abla5}{\textcolor{blue}{(e)}}). 
This indicates that designing an appropriate FFN is critical for the transformer block.

\textbf{B. Skip Connection and Fusion Block}. To validate the weighted connection and cross-layer attention fusion block, we conduct ablation studies by progressively removing the corresponding components: (1) skip, (2) $1\times1$ convolution, (3) skip with $1\times1$ convolution, (4) head CAFB, (5) tail CAFB, (6) all CAFBs. 
Table \ref{tab:component} shows the results in terms of PSNR and SSIM, which indicate that each component helps improve the results. The model improves significantly when including CAFB and weighted skip connections. We observe a minor gain when applying $1\times1$ convolutions.
  
\textbf{C. Wider {\it vs.} Deeper}. To understand the effect of width and depth in the network, we conduct ablation experiments to gradually increase the width (channels) and depth (number of encoder stages) of our LLFormer. Table \ref{tab:deep_wider} shows the results in terms of overhead, performance, and speed. The results demonstrate that LLFormer strikes the best tradeoff between performance and complexity (36.20/0.9867, 22.52G, 24.52M, 0.063s), compared to its wider or deeper counterparts. 

\section{Conclusion}
In this paper, we build the first large-scale low-light UHD image enhancement benchmark dataset, which consists of UHD-LOL4K and UHD-LOL8K subsets. Based on this dataset, we conduct comprehensive experiments for UHD-LLIE. To the best of our knowledge, this is the first attempt to specifically address the UHD-LLIE task. We propose the first transformer-based baseline network called LLFormer for UHD-LLIE. Extensive experiments show that  LLFormer significantly outperforms other state-of-the-art methods. The UHD-LOL dataset, together with LLFormer, will benefit the community in serving as a benchmark for LLIE and UHD-LLIE tasks.

\section{Acknowledgements}
This work was supported in part by the National Natural Science Foundation of China (Grant No. 61672273, 61832008), in part by Shenzhen Science and Technology Program (No. JSGG20220831093004008, JCYJ20220818102012025).

{
 \bibliography{aaai23}
 }
\end{document}